\documentclass[sigconf, 10pt]{acmart}

\usepackage[english]{babel}
\usepackage{blindtext}
\usepackage{caption}
\usepackage{subcaption}
\usepackage{color}
\usepackage{tabularx}
\usepackage{multirow}
\usepackage{multicol}
\usepackage{graphicx}
\usepackage{tabularray}
\usepackage{float}
\usepackage{placeins}
\usepackage{xcolor} 
\usepackage{todonotes}
\usepackage{array} 
\usepackage{makecell} 
\usepackage{booktabs} 
\usepackage{geometry}
\usepackage{microtype}
\renewcommand\footnotetextcopyrightpermission[1]{} 
\setcopyright{none}

\acmDOI{}

\acmISBN{}

\begin{document}

\title{HALO: A Heterogeneity-Aware Language-Aligned IMU Foundation Model for Open-Set Human Activity Recognition}

\author{Zihan Ding}
\affiliation{%
  \institution{Hong Kong University of Science and Technology}
  \city{Hong Kong}
  \country{China}
}
\email{alxd219p1@gmail.com}

\author{Liyu Zhang}
\affiliation{%
  \institution{Hong Kong University of Science and Technology}
  \city{Hong Kong}
  \country{China}
}
\email{lzhangcx@connect.ust.hk}

\author{Xiaomin Ouyang}
\affiliation{%
  \institution{Hong Kong University of Science and Technology}
  \city{Hong Kong}
  \country{China}
}
\email{xmouyang@cse.ust.hk}

\renewcommand{\shortauthors}{Ding, Zhang and Ouyang}

\begin{abstract}
Human Activity Recognition (HAR) using inertial measurement units (IMUs) enables a wide range of applications, yet the field still lacks a unified model that can generalize across diverse subjects, devices, and activities. Training such a model is difficult due to two key challenges: \emph{sensing heterogeneity}---differences in sampling rates, channel configurations, and sensor placements---and \emph{poor generalization} to unseen activities and label vocabularies.
We introduce HALO (\textbf{H}eterogeneity-\textbf{A}ware \textbf{L}anguage-aligned \textbf{O}pen-set model), a domain-specific IMU foundation model that addresses both challenges through a two-stage training framework\footnote{Preprint of work completed in March 2026, released as a citable record of that version; a substantially revised and extended version is in preparation. The source code and pretrained models for HALO will be released upon publication.}.
Stage~1 pretrains the IMU encoder with heterogeneity-aware self-supervised learning, including adaptive-pooling tokenization, channel-independent feature extraction, and \emph{contextualized sensor conditioning} that injects natural-language sensor descriptions into each channel embedding. Stage~2 aligns this IMU encoder with text embeddings via \emph{synonym-aware soft contrastive learning}, enabling open-set recognition via cosine-similarity retrieval without per-dataset classifiers.
Trained on 10 public HAR datasets and evaluated on 7 held-out datasets, HALO outperforms five state-of-the-art baselines on all 8 aggregate metrics, and still leads on 3 of 4 settings under baseline-matched inputs. Despite using only ${\sim}$35M trainable parameters---10$\times$ fewer than the latest foundation model MOMENT (341.2M)---HALO improves zero-shot open-set accuracy, measured over all 87 training labels, by 13.7 percentage points. On two further datasets with severe distribution shift, every model including HALO collapses zero-shot. A video demonstration of HALO's performance in real world is available at \url{https://youtu.be/rooVKragtFU}.

\end{abstract}

\maketitle

\section{Introduction}


Inertial measurement units (IMUs) are among the most widely deployed sensors in consumer electronics such as smartwatches and smartphones, enabling human activity recognition (HAR) for applications in health monitoring~\cite{mhealth}, rehabilitation~\cite{pamap2}, sports analytics~\cite{dsads}, and human--computer interaction. However, developing a unified model for HAR that generalizes across diverse subjects, devices, and activities remains challenging due to two fundamental issues. First, \emph{sensing heterogeneity}: datasets differ substantially in sampling rates (20--100 Hz), sensor suites (3--45 channels spanning accelerometers, gyroscopes, and magnetometers at various body placements), and label vocabularies with inconsistently overlapping activity names. Second, \emph{limited generalization}: models trained on one dataset typically transfer poorly to others, especially when faced with unseen activities or mismatched label spaces.

Addressing these challenges requires a model that can (i) operate at arbitrary sampling rates without architectural modifications, (ii) handle variable sensor suites and placements, and (iii) recognize activities absent from training---i.e., support open-vocabulary recognition under label shift. Existing approaches address these axes in isolation. Self-supervised methods (LiMU-BERT~\cite{limubert}, CrossHAR~\cite{crosshar}) improve transferability but require closed-world classifiers. General-purpose foundation models (MOMENT~\cite{moment}) cover diverse signal types but lack HAR-specific inductive biases. Language-aligned models (LanHAR~\cite{lanhar}) decouple the label interface via text similarity, but without strong sensor pretraining, naive text alignment proves insufficient for robust zero-shot transfer.

\begin{figure}[t]
    \centering
    \setlength{\abovecaptionskip}{2pt}
    \setlength{\belowcaptionskip}{-4pt}
  \includegraphics[width=\columnwidth]{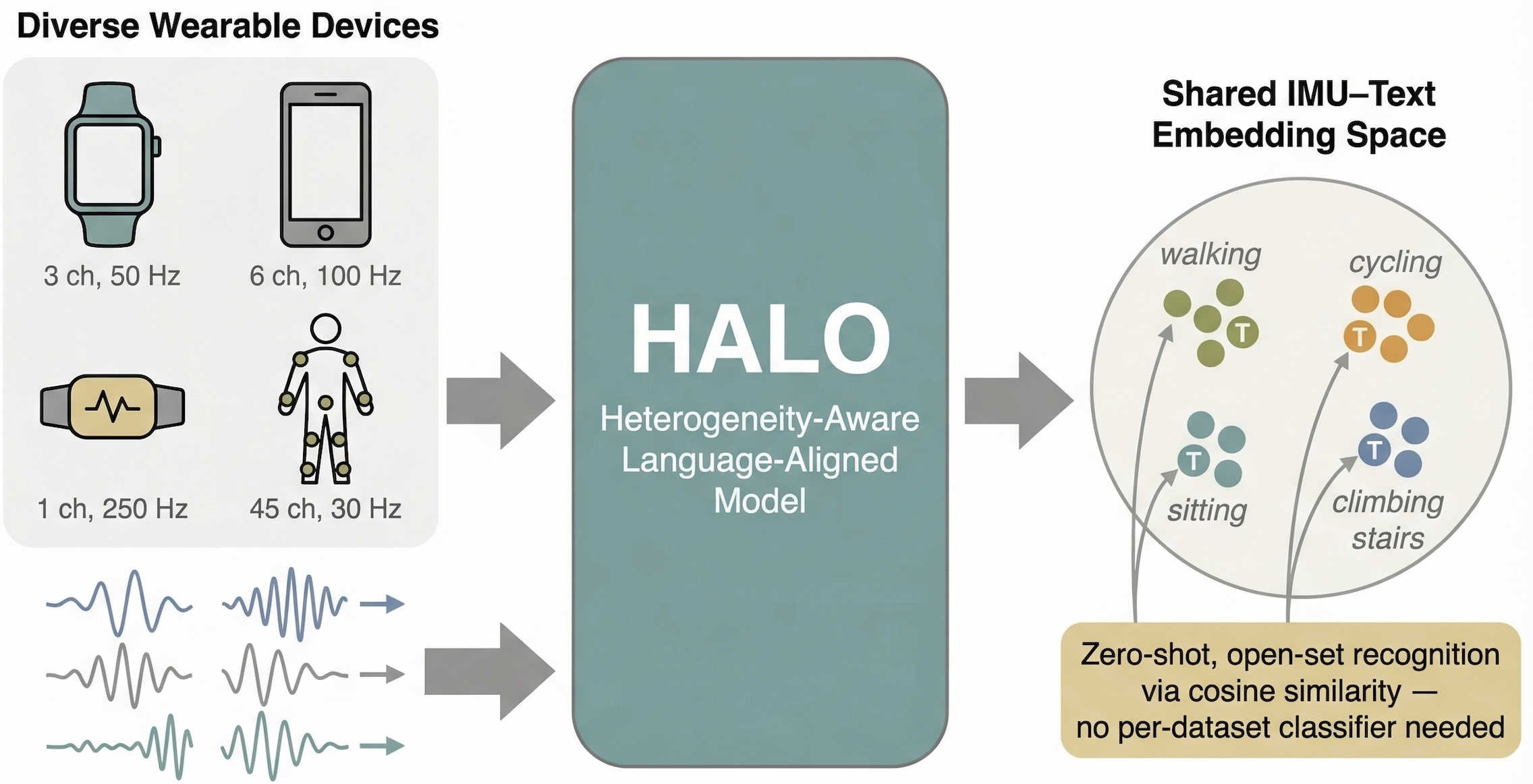}
  \caption{HALO is a domain-specific IMU foundation model that can tackle various sensing heterogeneity for open-set human activity recognition.
  }
  \label{fig:hook}
\end{figure}

We present HALO (\textbf{H}eterogeneity-\textbf{A}ware \textbf{L}anguage-aligned \textbf{O}pen-set model), a domain-specific IMU foundation model that enables a new deployment capability: a single pretrained model that accepts diverse wearable and phone IMU configurations---varying sampling rates, channel counts, and sensor placements---and recognizes activities via open-vocabulary cosine-similarity retrieval, with no per-dataset classifier or architectural modification required.
HALO achieves this through a two-stage framework driven by the three design requirements above.
Stage~1 pretrains a heterogeneity-aware IMU encoder via self-supervised objectives (masked patch autoencoding and contrastive learning), using (1) seconds-based adaptive-pooling tokenization for rate invariance, (2) channel-independent processing for variable sensor suites, and (3) \emph{contextualized sensor conditioning} that injects natural-language descriptions of each channel's sensor configuration (e.g., ``Accelerometer X-axis, waist-mounted phone'') to provide placement context.
Stage~2 aligns this encoder with a frozen SentenceBERT backbone~\cite{sbert} via \emph{synonym-aware soft contrastive learning}: soft targets derived from text--text similarity allow semantically related labels (e.g., ``walking'', ``strolling'', ``ambulating'') to reinforce one another, resolving the label-conflict problem inherent to multi-dataset training. At inference, activity recognition is cosine-similarity retrieval over any text label library---eliminating per-dataset classifiers.

We train HALO on 10 public HAR datasets and evaluate it on seven held-out test datasets---five main test datasets with high label coverage and two severe out-of-domain datasets containing entirely unseen activities---under four evaluation settings: zero-shot open-set (predicting from all 87 training labels with no knowledge of which labels appear in the test set), zero-shot closed-set (predicting only from the labels present in the test dataset), and 1\% and 10\% supervised fine-tuning. Against five state-of-the-art baselines spanning reconstruction-based, text-aligned, and LLM-based paradigms, HALO leads on all 8 aggregate metrics across the five main test datasets, improving zero-shot open-set accuracy by 13.7 percentage points over the general-purpose foundation model MOMENT~\cite{moment} (42.0\% vs.\ 28.3\%) while using only ${\sim}$35M trainable parameters---10$\times$ fewer than MOMENT (341.2M). On two severe out-of-domain datasets, all models---including HALO---collapse in zero-shot, confirming the limits of current cross-dataset transfer under extreme sensor or label shift; yet HALO adapts effectively with as little as 1\% labeled data, recovering to the strongest supervised performance on the harder of the two datasets.

We make the following contributions:
\begin{itemize}
  \item We systematically investigate sensing heterogeneity and open-set generalization as the two fundamental barriers preventing current IMU models from serving as general-purpose foundations for mobile HAR, and derive three concrete design requirements: rate-agnostic tokenization, variable-channel processing, and synonym-robust label alignment.
  \item We propose \textbf{HALO}, a domain-specific IMU foundation model that jointly addresses all three requirements through a two-stage training framework. The key integration insight is that heterogeneity-aware pretraining (adaptive-pooling tokenization, channel-independent processing, and \emph{contextualized sensor conditioning} via natural-language placement descriptions) provides a robust encoder that can then be aligned to text via \emph{synonym-aware soft contrastive learning}, enabling a deployment capability no prior system achieves: a single model that accepts diverse IMU configurations and recognizes activities via open-vocabulary cosine-similarity retrieval.
  \item We establish a unified cross-dataset benchmark spanning 10 training and 7 test datasets with four evaluation settings (eight metrics), and demonstrate that HALO outperforms five state-of-the-art baselines on all 8 aggregate metrics---including under baseline-equivalent input conditions (20\,Hz, generic descriptions)---with ${\sim}$35M trainable parameters. We will release the pretrained model checkpoint and codebase to facilitate reproducible research.
\end{itemize}

\section{Related Work}
\label{sec:related}

\textbf{IMU-based HAR.}
Deep learning for HAR has progressed from single-dataset CNN-LSTM models~\cite{deepconvlstm} to approaches addressing practical heterogeneity: Cosmo~\cite{cosmo} demonstrates contrastive fusion for small-data multimodal HAR, UniHAR~\cite{unihar} tackles cross-user and cross-device generalization via physics-informed augmentation, and ClusterFL~\cite{clusterfl} captures inter-user relationships through federated learning. Despite these advances, all remain architecturally bound to a fixed channel count, sampling rate, and closed-world label set, limiting transferability across sensing setups.

\textbf{Self-supervised pretraining and foundation models for IMU sensing.}
Self-supervised learning reduces dependence on labeled IMU data: Contrastive methods learn augmentation-invariant representations~\cite{simclr}, while masked autoencoding has been adapted from vision~\cite{mae} and language~\cite{bert} to IMU sequences. For example, LiMU-BERT~\cite{limubert} pretrains via span-masked reconstruction at 20\,Hz, and CrossHAR ~\cite{crosshar} combines masked reconstruction with contrastive learning for cross-dataset HAR. PatchTST~\cite{patchtst} demonstrates that patch-based channel-independent tokenization improves time-series modeling, a design principle we adopt. However, fixed-length positional encodings constrain variable sampling rates, and all methods require a dataset-specific classifier. Foundation models for general time-series data aim for broader coverage: MOMENT~\cite{moment} is a T5-style encoder pretrained on 13 time-series domains via masked patch reconstruction (341.2M parameters), and UniMTS~\cite{unimts} targets unified pretraining for motion time series across heterogeneous sensor configurations. In the tabular domain, Mitra~\cite{mitra} demonstrates that curated mixtures of synthetic priors can train foundation models that generalize across diverse real-world datasets via in-context learning. These general-purpose approaches trade domain-specific inductive biases for breadth; HALO occupies an intermediate position as a domain-specific foundation model for wearable sensing with HAR-specific designs and a language-based label interface.

\textbf{Language-grounded perception.}
CLIP~\cite{clip} demonstrated that aligning vision and text in a shared embedding space enables open-vocabulary zero-shot classification without per-class classifiers. ImageBind~\cite{imagebind} extends this to multiple modalities including IMU, but is not designed for cross-dataset HAR or variable sensor configurations. LanHAR~\cite{lanhar} applies CLIP-style alignment to HAR by training a sensor Transformer against LLM-generated activity descriptions. LLaSA~\cite{llasa} takes an alternative approach, leveraging LLMs directly for natural language reasoning about IMU data. Unlike LanHAR, which lacks self-supervised pretraining and assumes fixed sensor configurations, HALO integrates heterogeneity-aware pretraining with synonym-aware contrastive alignment, enabling a capability none of these prior systems achieve: a single deployable model that handles variable sampling rates, channel counts, and sensor placements for open-set HAR.

\section{Motivation}
\label{sec:motivation}

We identify two fundamental challenges that prevent current IMU-based HAR models from serving as general-purpose foundations: \emph{sensing heterogeneity} across devices and deployments, and \emph{limited generalization} to unseen activity distributions. We analyze each challenge through the lens of real-world applications and derive the design requirements that motivate HALO.

\subsection{Sensing Heterogeneity Across Devices and Deployments}

\begin{figure}[t]
    \centering
    \setlength{\abovecaptionskip}{2pt}
    \setlength{\belowcaptionskip}{-4pt}
  \includegraphics[width=\columnwidth]{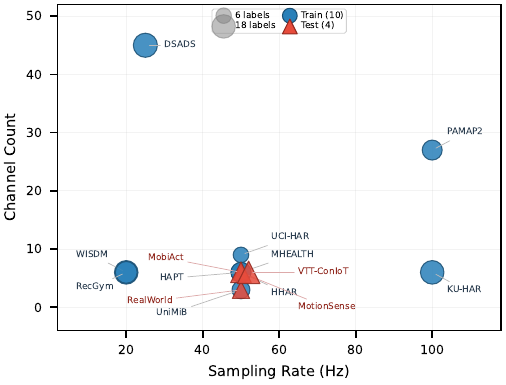}
  \caption{Dataset heterogeneity across 14 HAR datasets in our benchmark. Training datasets (blue circles) and test datasets (red triangles) span 20--100\,Hz in sampling rates, 3--45 channels, and 4--19 activity labels.}
  \label{fig:dataset_scatter}
\end{figure}

Wearable and mobile devices differ widely in their sensing configurations, leading to substantial data heterogeneity across HAR datasets. For example, a smartwatch with a 3-axis accelerometer at 50\,Hz produces significantly different raw data than a multi-sensor body suit capturing 51 channels at 100\,Hz, even when both record the same physical activity. 
To systematically analyze such heterogeneity, we analyze the 17 datasets included in Table~\ref{tab:datasets} of our benchmark. Figure~\ref{fig:dataset_scatter} visualizes the distributions of the key characteristics of these datasets.


\textbf{Sampling rate variation.}
Sampling rates range from 20\,Hz to 100\,Hz (Figure~\ref{fig:dataset_scatter}), so a 4-second window contains 80 to 400 timesteps depending on the device. A fixed positional encoding tied to timestep index conflates physical time with sequence length, causing identical motions at different rates to appear structurally different.

\textbf{Channel and placement diversity.}
Channel counts range from 3-axis wrist accelerometers to 51-channel multi-IMU body suits. A fixed-width input layer cannot accommodate this variability without padding artifacts or dataset-specific input heads.


\subsection{Limited Generalization to New Distributions and Activities}

Beyond sensing heterogeneity, existing approaches share a structural limitation: they produce representations consumed by a fixed, closed-world classifier. This is adequate within a single dataset but fails in two important transfer scenarios.

\textbf{Label vocabulary inconsistency.} Label vocabularies differ in both naming and granularity: ``walking,'' ``level walking,'' ``walking on a flat surface,'' and ``normal walk'' all denote the same activity across different datasets, while ``sport'' and ``rope skipping'' represent very different levels of specificity. Because each research group defines its own taxonomy independently, standard one-hot supervision treats ``jogging'' and ``running'' as maximally different, producing contradictory gradients when both labels appear in the training set for similar motions.

The first is \textit{label shift}: a model trained on ten datasets with a combined 87 unique activity labels is never exposed to the specific label strings used in a held-out test set, even when the underlying motions overlap. For classifier-based models, this means the output vocabulary cannot cover the test labels, and synonym mismatches (e.g., predicting ``jogging'' when the ground truth is ``running'') count as errors even when semantically correct. The second is \textit{genuinely novel activities}: our test datasets include activities with no semantic analog in the training label set (e.g., \textit{car\_step\_in} in MobiAct, industrial activities in VTT-ConIoT), making correct predictions structurally impossible for any closed-world model.

Text-aligned models can avoid the closed-world limitation: inference is cosine-similarity retrieval over an arbitrary text library, so the candidate set can be changed at runtime without retraining. However, effective text alignment requires both (i) a sensor encoder robust to the heterogeneity described above, and (ii) a training objective that handles synonym variation in label strings rather than treating them as distinct classes.

\subsection{Design Requirements}

The analysis above yields three concrete requirements for a cross-device IMU model: (R1)~the model must handle arbitrary sampling rates, channel counts, and sensor placements without dataset-specific architectural changes; (R2)~it must support open-vocabulary recognition over arbitrary label sets at test time, including labels not seen during training; and (R3)~it must align semantically equivalent labels during training to avoid contradictory gradients from synonym variation. HALO achieves R1 through heterogeneity-aware pretraining (Section~\ref{sec:design}), and R2--R3 through synonym-aware IMU-to-text contrastive alignment (Section~\ref{sec:design}).

\section{System Overview}
\label{sec:overview}

\begin{figure}[t]
    \centering
    \setlength{\abovecaptionskip}{2pt}
    \setlength{\belowcaptionskip}{-4pt}
  \includegraphics[width=\columnwidth]{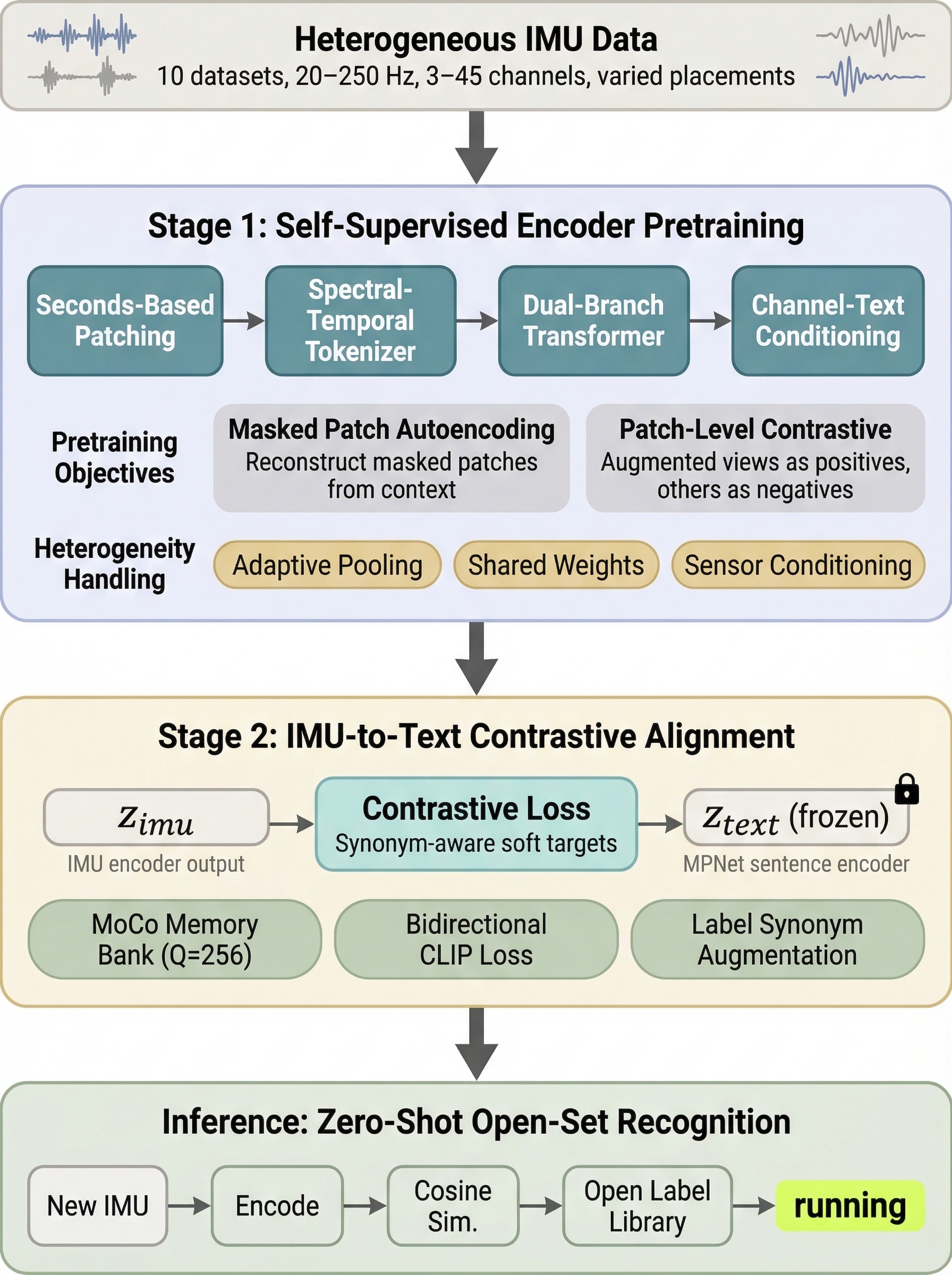}
  \caption{System overview of HALO. The sensing heterogeneity is addressed through heterogeneity-aware pretraining of the IMU encoder (Stage~1), followed by synonym-aware soft contrastive learning to align sensor features with text semantics (Stage~2), enabling generalizable and open-set activity recognition. Zero-shot inference is achieved by swapping label banks at runtime without retraining.  }
  \label{fig:overview}
\end{figure}

HALO is an IMU-to-text semantic alignment model trained across heterogeneous HAR datasets for open-set, zero-shot activity recognition. The system addresses the design requirements identified in Section~\ref{sec:motivation} through a two-stage training framework. Figure~\ref{fig:overview} illustrates the end-to-end pipeline.

\textbf{Stage~1: Heterogeneity-aware pretraining.} The IMU encoder maps variable-rate, variable-channel sessions to a fixed embedding $\mathbf{z}_{\text{imu}} \in \mathbb{R}^{D}$ through adaptive-pooling tokenization, channel-independent processing, and contextualized sensor conditioning, then is pretrained via masked autoencoding and patch-level contrastive learning (Section~\ref{sec:heterogeneity}).

\textbf{Stage~2: Synonym-aware IMU-to-text alignment.} The pretrained encoder is aligned with a frozen Sentence-BERT text encoder via synonym-aware soft contrastive learning that derives soft target distributions from text--text similarity, resolving label conflicts across datasets and mapping both modalities into a shared 384-dimensional space (Section~\ref{sec:alignment_design}).

\textbf{Inference.} At test time, recognition is cosine-similarity retrieval: given $\mathbf{z}_{\text{imu}}$, we select the label embedding with maximum similarity, supporting arbitrary candidate inventories including open-vocabulary settings. No per-dataset classifier is needed. The full model contains ${\sim}$35M trainable parameters (8-layer dual-branch IMU encoder ${\sim}$19M, alignment and projection heads ${\sim}$16M) plus a single frozen Sentence-BERT backbone (all-MiniLM-L6-v2, 22.7M) shared for both channel conditioning and text alignment. At deployment, the sensor description string is specified once when the app is configured; the corresponding text embeddings are computed once and cached, adding zero per-inference cost. Likewise, all candidate label embeddings can be precomputed offline.


\section{Design of HALO}
\label{sec:design}

This section details HALO's two stages: the hetero\-geneity-aware encoder with self-supervised pretraining (Stage~1, Section~\ref{sec:heterogeneity}) and synonym-aware contrastive text alignment (Stage~2, Section~\ref{sec:alignment_design}).

\subsection{Heterogeneity-aware Pretraining}
\label{sec:heterogeneity}

\begin{figure}
    \centering
    \setlength{\abovecaptionskip}{2pt}
    \setlength{\belowcaptionskip}{-4pt}
  \includegraphics[width=\columnwidth]{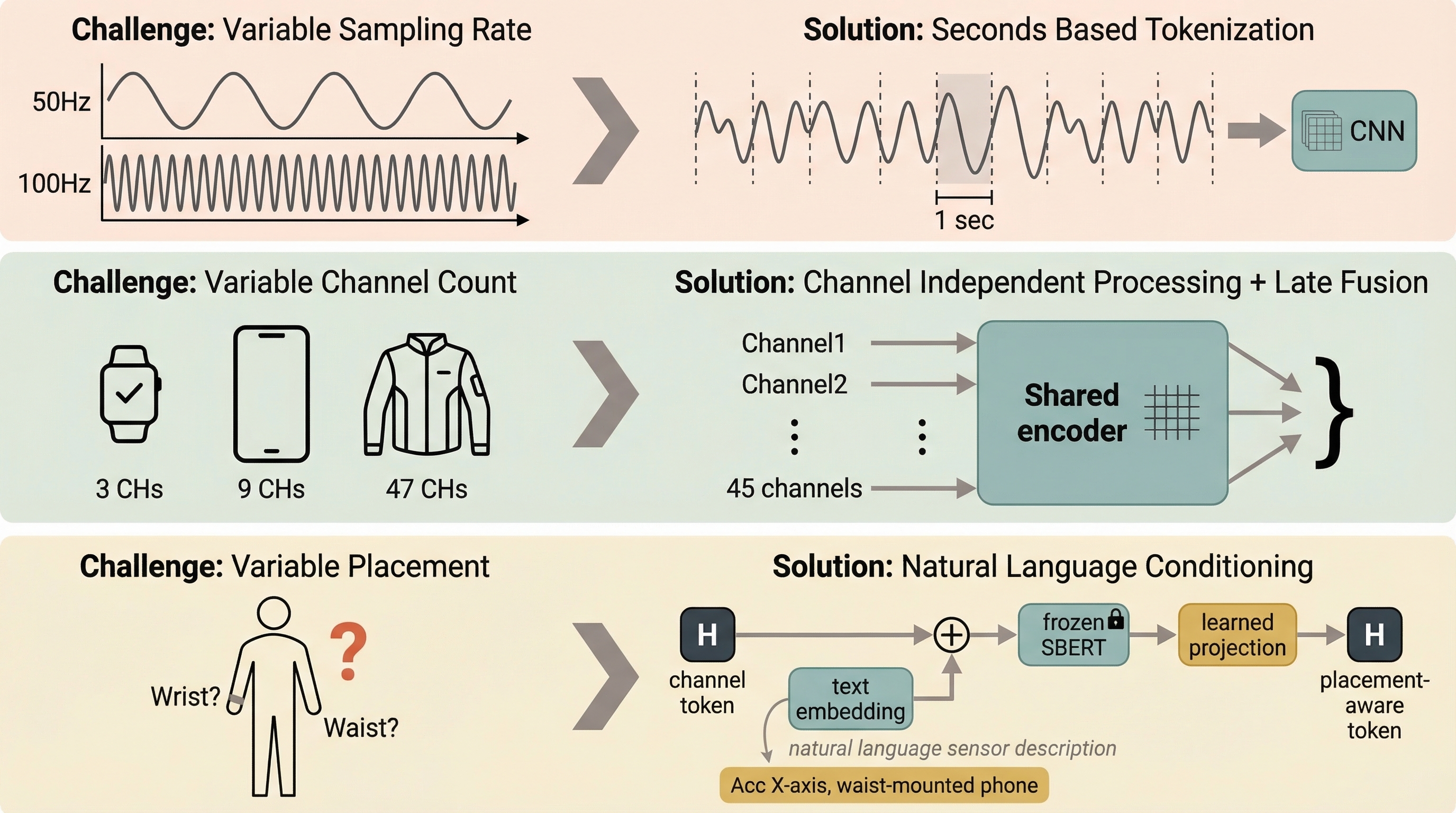}
  \caption{Heterogeneity-aware Pretraining. Each row shows a challenge (left) and the corresponding design in HALO (right): adaptive-pooling tokenization for variable sampling rates; channel-independent feature extraction for variable channels; and contextualized sensor conditioning for variable sensor placements.}
  \label{fig:encoder}
\end{figure}

The encoder addresses the three heterogeneity challenges identified in Section~\ref{sec:motivation} through dedicated components (Figure~\ref{fig:encoder}). At inference time, a sample flows through them in sequence: tokenization produces per-channel tokens $\mathbf{H}_0$, sensor conditioning enriches them to $\mathbf{H}'$, and the channel-independent encoder fuses them into the session embedding $\mathbf{z}_{\text{imu}}$. We present each component below, grouping the encoder with its fusion mechanism before describing conditioning, since the encoder architecture is easier to understand as a self-contained unit.

\subsubsection{Adaptive-Pooling Tokenization for Variable Sampling Rates}
\label{sec:tokenization}

HAR datasets are collected at different native sampling rates---for example, PAMAP2~\cite{pamap2} records at 100\,Hz while Shoaib~\cite{shoaib} uses 50\,Hz and WISDM~\cite{wisdm} operates at 20\,Hz. A conventional fixed-length tokenizer that expects $L$ samples per patch would either require resampling every dataset to a common rate (distorting high-frequency content or fabricating samples) or training separate models per rate. Rather than resampling, HALO segments sessions in \emph{seconds} rather than samples, so every patch spans the same physical duration regardless of rate. The resulting variable-length patches are then mapped to fixed-dimension tokens via adaptive pooling, decoupling the entire encoder from any specific sampling rate.

\textbf{Seconds-based patching.}
Each session is segmented into fixed-duration patches (dataset-specific, typically 0.75--2.0\,s), yielding $\mathbf{X}\in\mathbb{R}^{B\times P\times L\times C}$ where $L$ is the native patch length and $C$ the channel count. A 1-second patch at 50\,Hz contains $L{=}50$ samples while the same patch at 100\,Hz contains $L{=}100$; the variable $L$ is absorbed downstream by adaptive pooling rather than resampling.
During training, patch durations are randomly sampled from a dataset-specific range (e.g., 0.75--1.25\,s) as a temporal augmentation; augmentation is disabled at evaluation.

\textbf{Per-patch, per-channel normalization.}
Each patch-channel pair is independently z-score normalized:
\[
\hat{\mathbf{X}}_{b,p,:,c}=\frac{\mathbf{X}_{b,p,:,c}-\mu_{b,p,c}}{\sigma_{b,p,c}+\epsilon},
\]
where $\mu_{b,p,c}$ and $\sigma_{b,p,c}$ are computed over the $L$ timesteps of patch $(b,p)$ channel $c$. This local normalization removes device- and subject-specific amplitude biases without requiring global statistics.

\textbf{Spectral-temporal tokenizer.}
\label{sec:architecture}
Each normalized patch-channel pair is converted into a $D$-di\-men\-sion\-al token ($D{=}384$) via a hybrid feature extractor with shared weights across channels. The temporal branch applies a multi-scale 1-D CNN (kernel sizes 3, 5, 7) followed by adaptive average pooling to produce a fixed-length output regardless of~$L$; the spectral branch computes a fixed-$n$ FFT magnitude spectrum and projects it through a learned MLP\@. Combining both branches captures complementary time-domain shape and frequency-domain periodicity cues. The outputs are concatenated to form the initial token tensor $\mathbf{H}_0\in\mathbb{R}^{B\times P\times C\times D}$.\looseness=-1

Because both branches use adaptive pooling, the tokenizer natively accepts any sampling rate without resampling or architectural changes---a single trained model can process 20\,Hz smartphone data and 100\,Hz clinical IMU data through the same weights. This is a key advantage over fixed-length tokenizers such as PatchTST~\cite{patchtst}, which require a predetermined number of samples per patch and therefore either resample all inputs to a common rate (distorting high-frequency content at downsampled rates and fabricating samples at upsampled rates) or maintain separate tokenizer configurations per dataset. Patch-size sensitivity analysis (Section~\ref{sec:patch_sensitivity}) confirms that the fixed 1.0\,s default is within 1.3\,pp of per-dataset optima across the full 20--100\,Hz range.

\subsubsection{Channel-Independent Feature Extraction for Variable Channel Counts}
\label{sec:encoder_fusion}

Different datasets expose vastly different sensor suites: WISDM~\cite{wisdm} provides a single 3-axis accelerometer ($C{=}3$), while DSADS~\cite{dsads} uses 45 channels from accelerometers, gyroscopes, and magnetometers at five body locations. A fixed-width architecture would either waste capacity on small datasets or fail entirely on larger ones. To handle this, HALO treats every channel identically: the tokenizer, Transformer, and fusion modules all share parameters across channels, and variable-length channel dimensions are zero-padded with binary masks propagated to every attention and pooling layer so that padded channels never contribute to computation.

\textbf{Dual-branch Transformer.}
The token tensor---after sensor conditioning (Section~\ref{sec:conditioning}, below)---is processed by $L_e$ Transformer layers ($L_e{=}8$), each alternating two attention stages: (1)~\emph{temporal self-attention} over patches within each channel (reshape to $(BC,P,D)$), which captures how each individual sensor signal evolves over time, and (2)~\emph{cross-channel self-attention} over channels within each patch (reshape to $(BP,C,D)$, padded channels masked), which captures correlations between different sensors at the same time step. Each stage is followed by a shared FFN, yielding $\mathbf{H}\in\mathbb{R}^{B\times P\times C\times D}$. Sinusoidal positional encodings~\cite{transformer} with a learnable scale parameter inject temporal order over the patch dimension. The alternating design allows the model to first build per-sensor temporal representations and then exchange information across sensors at every layer, progressively building richer multi-sensor features.

\textbf{Cross-channel fusion.}
For each patch, a small set of learnable queries attends over the $C$ channel tokens via multi-query attention, producing a fused patch sequence $\mathbf{U}\in\mathbb{R}^{B\times P\times D}$. Channel masks prevent padded channels from influencing attention. This reduces the variable-width channel dimension to a fixed-width representation, enabling downstream modules to operate regardless of the original sensor count.

\textbf{Temporal pooling and projection.}
A lightweight Transformer over $\mathbf{U}$ followed by a second multi-query attention module pools the patch axis to yield $\mathbf{v}\in\mathbb{R}^{B\times D}$, which an MLP projection head maps to the session embedding $\mathbf{z}_{\text{imu}}\in\mathbb{R}^{B\times D_s}$ ($D_s{=}384$, $\ell_2$-normalized). HALO also supports \emph{per-patch prediction}: each fused patch token $\mathbf{U}_{b,p}$ is independently projected and scored against text labels via cosine similarity, with per-patch predictions aggregated by majority vote. All zero-shot results in this paper use per-patch majority voting; supervised fine-tuning uses the pooled global embedding $\mathbf{z}_{\text{imu}}$.

The channel-independent design means the same trained weights handle 3-channel smartphone data and 45-channel full-body sensor suites with no architectural changes across all 17~datasets. At training time, channels from different datasets are mixed within each batch, exposing the encoder to diverse sensor types and body locations in every gradient step. The shared-parameter approach also acts as an implicit regularizer: the model cannot memorize channel-specific shortcuts, encouraging generalizable motion features that transfer across sensor configurations.\looseness=-1

\subsubsection{Contextualized Sensor Conditioning for Variable Sensor Placements}
\label{sec:conditioning}

Because the tokenizer shares weights across all channels, the tokens $\mathbf{H}_0$ carry temporal and spectral information but are agnostic to \emph{where} on the body each channel was recorded or \emph{what} type of sensor produced it. A wrist accelerometer X-axis and a waist gyroscope Z-axis may produce similar waveforms for different activities---for example, arm-swing acceleration during walking looks similar from a wrist accelerometer and a hip accelerometer, yet the two carry different biomechanical meaning. Without placement context, the encoder cannot disambiguate them. HALO addresses this by injecting sensor placement and modality context into each channel token via natural-language descriptions, using the same frozen text encoder that will later be used for activity-label alignment. This converts the heterogeneity problem into a representational advantage: the model learns that ``Accelerometer X-axis, waist-mounted smartphone'' and ``Accelerometer X-axis, wrist-mounted smartwatch'' are related but distinct, leveraging the semantic structure of natural language.

Concretely, each channel token $\mathbf{H}_{0,b,p,c}$ is conditioned on a per-channel semantic description (e.g., ``Accelerometer X-axis, waist-mounted smartphone'').
A frozen Sentence-BERT backbone~\cite{sbert} (all-MiniLM-L6-v2, 384-dim)---shared with the text encoder used in Stage~2 alignment---encodes the description, and a learned two-layer residual projection with ReLU maps it to $\mathbf{e}_c \in \mathbb{R}^D$.
The conditioned token is:
$\mathbf{H}'_{b,p,c} = \mathbf{H}_{0,b,p,c} + \gamma\,\mathbf{e}_c$,
where $\gamma$ is a learnable scale initialized to~0.1 so that conditioning acts as a gentle bias early in training and grows as the model learns to exploit it.\looseness=-1{}
The conditioned tokens $\mathbf{H}'$ are then passed to the dual-branch Transformer (Section~\ref{sec:encoder_fusion}).

Descriptions are constructed per dataset from a simple template combining modality, axis, and placement metadata (e.g., ``Accelerometer X-axis, waist-mounted smartphone''). Each dataset's channel descriptions are specified once during data preparation and remain fixed throughout training and evaluation. For datasets without placement annotations (e.g., UCI-HAR provides only ``smartphone'' without specifying pocket vs.\ hand), we use the available metadata; datasets with no sensor metadata fall back to generic descriptions (e.g., ``IMU channel''). At inference, the model leverages rich descriptions when available and falls back to generic ones otherwise. Sensor conditioning is the single largest contributor to zero-shot transfer ($-$19.6\,pp ZS-O when removed; Table~\ref{tab:component_ablation}); even with generic descriptions HALO outperforms baselines on 3 of~4 settings (Table~\ref{tab:fairness_control}).\looseness=-1

\subsubsection{Self-supervised Pretraining}
\label{sec:pretraining}

Before alignment with text, the encoder must learn general-purpose IMU repre\-sen\-ta\-tions from unlabeled data across all 10 training datasets. The difficulty is that heterogeneous sensor configurations and activities can cause the encoder to overfit to dataset-specific artifacts (e.g., amplitude ranges, noise profiles) rather than learning transferable motion patterns. Stage~1 addresses this by training the encoder end-to-end with two complementary self-supervised objectives---masked autoencoding for local signal reconstruction and patch-level contrastive learning for global discriminability---combined with signal augmentation that forces the model to generalize across noise conditions.

\textbf{Masked autoencoding.} A fixed fraction of patches (mask ratio 0.5) are replaced by a learned mask token, and the temporal encoder reconstructs the normalized input signal at masked positions:
\[
\mathcal{L}_{\text{mae}}=\frac{1}{|\mathcal{M}|}\sum_{(b,p)\in\mathcal{M}} \left\| \hat{\mathbf{u}}_{b,p} - \mathbf{u}_{b,p} \right\|_2^2.
\]
This objective encourages the encoder to capture fine-grained temporal patterns within each channel, such as the periodic swing phase in walking or the abrupt deceleration in sitting down. The 50\% mask ratio is chosen to balance reconstruction difficulty: lower ratios make the task too easy (the encoder can interpolate from nearby unmasked patches), while higher ratios deprive the model of sufficient context for meaningful reconstruction.

\textbf{Patch-level contrastive learning.} Two augmented views of each patch (jittering, scaling, time warping) are treated as positives and other in-batch patches as negatives under an InfoNCE objective~\cite{infonce} with temperature 0.2. This objective encourages patches from similar motions to cluster in embedding space while pushing dissimilar motions apart, building discriminative structure before any labels are introduced. Both objectives are combined with equal weights ($\lambda_{\text{mae}}=\lambda_{\text{con}}=1.0$).

Together, the dual objectives produce an encoder that captures both local signal fidelity (via MAE) and global motion discriminability (via contrastive learning). Signal augmentation is critical: it closes a 9.5\,pp generalization gap invisible on in-distribution validation but evident on held-out datasets (Section~\ref{sec:component_ablation}).

\subsection{Synonym-aware IMU-to-text Alignment}
\label{sec:alignment_design}

A key challenge in multi-dataset IMU-to-text alignment is \emph{label synonym conflict}: ``walking,'' ``strolling,'' and ``ambulating'' describe the same motion, yet standard contrastive objectives treat them as unrelated negatives, producing contradictory gradients when both appear in a batch. HALO resolves this through synonym-aware label augmentation (Section~\ref{sec:label_aug}) and soft contrastive alignment (Section~\ref{sec:soft_contrastive}).

\begin{figure}[t]
    \centering
    \setlength{\abovecaptionskip}{2pt}
    \setlength{\belowcaptionskip}{-4pt}
  \includegraphics[width=\columnwidth]{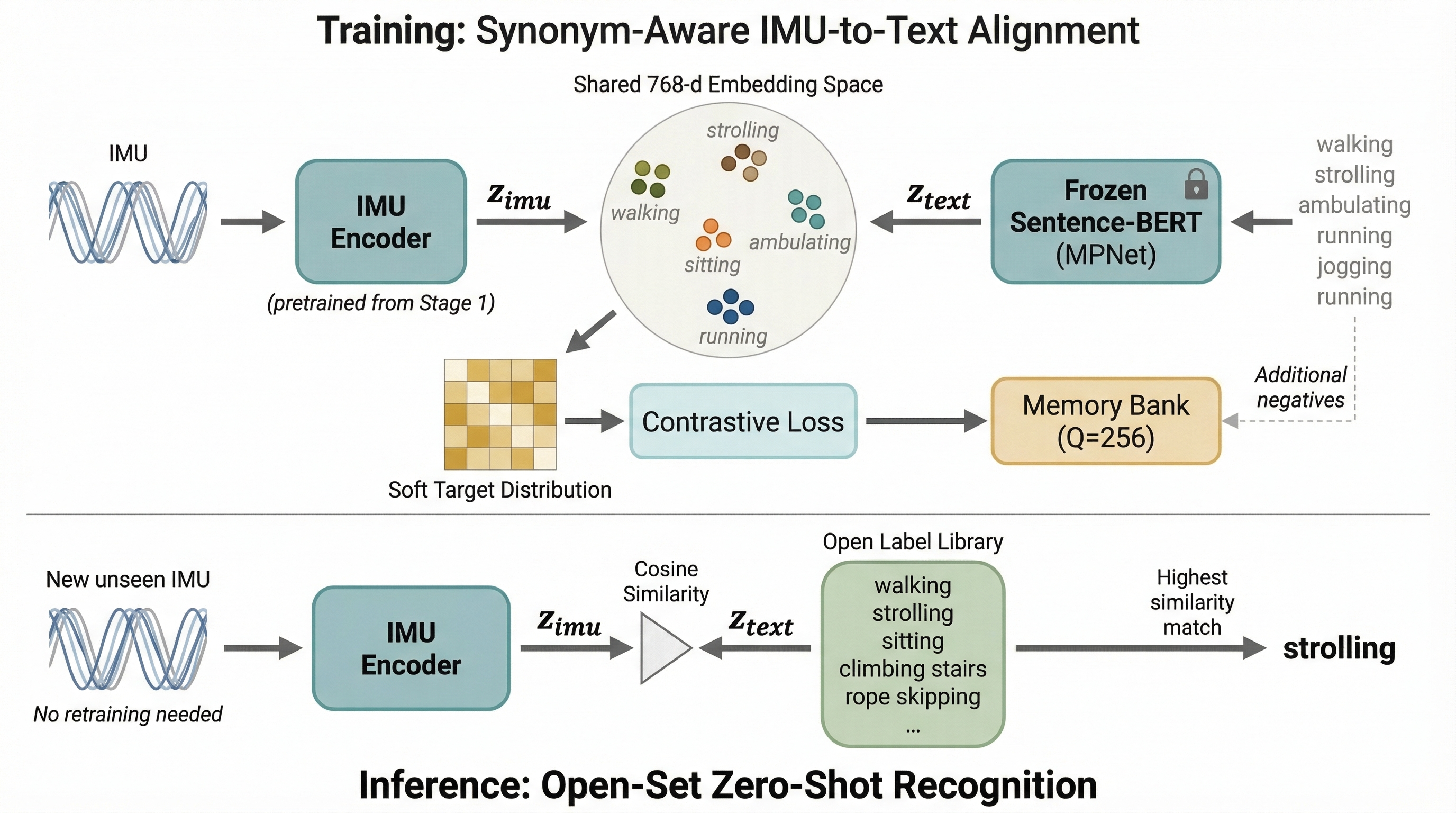}
  \caption{IMU-to-text alignment and zero-shot inference. \emph{Top:} During training, the IMU encoder is aligned with a frozen text encoder in a shared embedding space via synonym-aware contrastive learning, so that semantically related activities are clustered together. \emph{Bottom:} At inference, the IMU segment is encoded and matched against an \emph{open} label library via cosine similarity-enabling zero-shot recognition without retraining or dataset-specific classifiers.}
  \label{fig:alignment}
\end{figure}

\subsubsection{Synonym-Aware Label Augmentation}
\label{sec:label_aug}

Each dataset defines a curated synonym map (e.g., \textit{walking} $\leftrightarrow$ \textit{strolling} $\leftrightarrow$ \textit{ambulating}) and natural-language templates (e.g., ``person \{\}'', ``a person is \{\}''). Synonym maps are constructed once per dataset from WordNet and common HAR naming conventions, averaging 2.3~synonyms per activity across the 87-label vocabulary. During each Stage~2 step, with probability~0.8 the original label is replaced by a randomly sampled synonym wrapped in a random template (disabled at evaluation), diversifying the text embeddings seen for each activity and preventing the text encoder from overfitting to exact label strings in any single dataset.\looseness=-1

Such label augmentation reduces synonym conflict frequency in each batch ($-$7.0\,pp at 1\% when removed; Table~\ref{tab:component_ablation}), but cannot eliminate it entirely---residual synonyms still appear as negatives---motivating the soft targets described next.

\subsubsection{IMU-to-Text Alignment with Soft Contrastive Learning}
\label{sec:soft_contrastive}

Standard CLIP-style learning uses hard one-hot targets, forcing synonymous labels apart. HALO replaces these with \emph{soft} target distributions derived from text--text similarity, so synonymous labels receive partial credit---converting synonym conflict from a source of gradient noise into a learning signal that reinforces semantic structure.

\textbf{Text encoder and contrastive objective.}
A frozen Sentence-BERT backbone~\cite{sbert} (all-MiniLM-L6-v2, 384-dim)---the same model used for sensor conditioning---encodes augmented activity labels; learnable query-based pooling over its token outputs produces $\mathbf{z}_{\text{text}}\in\mathbb{R}^{B\times D_s}$.
Stage~2 aligns $\mathbf{z}_{\text{imu}}$ with $\mathbf{z}_{\text{text}}$ via a CLIP-style bidirectional contrastive objective. Because our batch sizes are limited by GPU memory (micro-batch 32 with gradient accumulation), we maintain FIFO queues~\cite{moco} of size $Q=256$ for both modalities to increase the effective number of negatives per step from 32 to 288, pre-filling them with a warmup pass before training begins. Similarity logits are computed as
\[
\mathbf{S}_{i\rightarrow t} = \alpha \, \mathbf{Z}_{\text{imu}}\, \mathbf{A}_{\text{text}}^\top, \quad
\mathbf{S}_{t\rightarrow i} = \alpha \, \mathbf{Z}_{\text{text}}\, \mathbf{A}_{\text{imu}}^\top,
\]
where $\mathbf{A}$ denotes the concatenation of the current batch and queue, and $\alpha=\exp(\texttt{logit\_scale})$ is a learnable temperature (CLIP-initialized, clamped to $[1,50]$).

\textbf{Adaptive soft targets.}
Raw text--text similarities cluster in a narrow range (0.4--0.9), producing near-uniform targets that collapse the embedding space. We apply \emph{adaptive sharpening}: z-score-normalize the in-batch similarity matrix $\mathbf{M}=\mathbf{Z}_{\text{text}}\mathbf{Z}_{\text{text}}^\top$, rescale with temperature $\tau_s{=}0.5$, and softmax to obtain sharpened distributions $\mathbf{T}$ (e.g., ``walking''/``strolling'' share 0.4 of target mass while ``walking''/``cycling'' share $<$0.01). Queue entries serve as hard negatives. The blended target $\mathbf{P}$ combines soft in-batch targets $\mathbf{T}$ with hard one-hot targets for queue entries. The bidirectional loss is:
\begin{align*}
\mathcal{L}_{i\rightarrow t} &= -\frac{1}{B}\sum_{b,j} \mathbf{P}_{b,j}\log\mathrm{softmax}(\mathbf{S}_{i\rightarrow t})_{b,j}, \\
\mathcal{L} &= \tfrac{1}{2}\bigl(\mathcal{L}_{i\rightarrow t}+\mathcal{L}_{t\rightarrow i}\bigr).
\end{align*}

\textbf{Zero-shot inference.}
At test time, the predicted activity is $\hat{y}=\arg\max_k \cos(\mathbf{z}_{\text{imu}}, \mathbf{z}_{\text{text}}^{(k)})$ over an arbitrary candidate label set. Swapping the label library at runtime enables open-set recognition without retraining (Section~\ref{sec:overview}).

Embedding-space analysis confirms 76.7\% nearest-neighbor accuracy between IMU and text centroids (Section~\ref{sec:embedding_analysis}).

\section{Experiments}
\label{sec:experiments}

\subsection{Methodology}
\label{sec:methodology}

We evaluate whether HALO supports (i)~robust performance under heterogeneous multi-dataset training, (ii)~open-set and closed-set zero-shot generalization to unseen datasets, and (iii)~sample-efficient supervised transfer. 

\begin{table}[t]
  \centering
  \resizebox{\columnwidth}{!}{
  \begin{tabular}{lcccc p{0.18\columnwidth} cc}
    \toprule
    Dataset & Hz & \#Subj & \#Act & \#Ch & Description & Cov. & Diff. \\
    \midrule
    UCI-HAR~\cite{ucihar}        & 50  & 30 &  6 & 9   & Smartphone IMU & & \\
    HHAR~\cite{hhar}             & ${\sim}$50 & 9 & 6 & 6 & Phone + watch & & \\
    UniMiB~\cite{unimibi}        & 50  & 30 & 17 & 3   & Phone acc, ADLs & & \\
    MHEALTH~\cite{mhealth}       & 50  & 10 & 12 & 6+  & Body IMU suite & & \\
    PAMAP2~\cite{pamap2}         & 100 &  9 & 18 & 27  & Multi-IMU & & \\
    WISDM~\cite{wisdm}           & 20  & 51 & 18 & 6   & Phone + watch & & \\
    DSADS~\cite{dsads}           & 25  &  8 & 19 & 45  & 5 body positions & & \\
    HAPT~\cite{hapt}             & 50  & 30 & 12 & 6   & Phone, transitions & & \\
    KU-HAR~\cite{kuhar}          & 100 & 90 & 18 & 6   & Phone, 18 acts & & \\
    RecGym~\cite{recgym}         & 20  & 10 & 12 & 6   & Watch, gym & & \\
    \midrule
    MotionSense$^\dagger$~\cite{motionsense} & 50 & 24 &  6 & 6 & Phone, basic & 100\% & Easy \\
    RealWorld$^\dagger$~\cite{realworld}     & 50 & 15 &  8 & 3 & Phone, acc-only & 100\% & Med \\
    MobiAct$^\dagger$~\cite{mobiact}         & 50 & 66 & 13 & 6 & Falls + ADLs & 85\% & Hard \\
    Shoaib$^\dagger$~\cite{shoaib}           & 50 & 10 &  7 & 6 & Multi-placement & 100\% & Med \\
    Opportunity$^\dagger$~\cite{opportunity}  & 30 &  4 &  4 & 6 & XSens body-worn & 100\% & Med \\
    HARTH$^\ddagger$~\cite{harth}            & 50 & 22 & 12 & 6 & Back/thigh acc & 100\% & Severe \\
    VTT-ConIoT$^\ddagger$~\cite{vttconiot}   & 52 & 12 & 16 & 6 & Construction & 50\% & Severe \\
    \bottomrule
  \end{tabular}}
  \caption{Dataset overview. Top: 10 training datasets. Bottom: 7 held-out test datasets.
  $\dagger$\,Main test datasets. $\ddagger$\,Severe OOD datasets; reported separately.
  \emph{Cov.}: Fraction of test activities that have an equivalent in the training set.
  \emph{Diff.}: Difficulty level determined by distribution shift and label novelty. }
  \label{tab:datasets}
   \vspace{-1em}
\end{table}

\textbf{Datasets.}
\label{sec:datasets}
We train jointly on ten public HAR datasets (Table~\ref{tab:datasets}). All HAR-pretrained models use the same 10 training datasets; MOMENT is pretrained on general time-series data; LLaSA uses published weights whose training data overlaps with two of our test sets (MotionSense, Shoaib).
We evaluate on seven held-out datasets \emph{never seen during any model's training}: five main test datasets (85--100\% label coverage) and two severe out-of-domain datasets reported separately.
HALO evaluates at native rates and LanHAR at 50\,Hz; the remaining baselines use 20\,Hz resampled data via the LIMU-BERT pipeline (Section~\ref{sec:eval_protocol}).

\textbf{Baselines.}
\label{sec:baselines}
We compare HALO against five baselines representing different pretraining paradigms (Table~\ref{tab:baselines}):

\emph{Text-aligned models} (HALO, LanHAR) predict via cosine similarity between sensor and text embeddings, enabling arbitrary label strings at test time (evaluated by exact string match). LanHAR fine-tunes SciBERT on activity text, then aligns a sensor Transformer.
\emph{Classifier-based models} (LiMU-BERT, MOMENT, CrossHAR) train a fixed 87-class classifier over the global training labels. LiMU-BERT uses span-masked reconstruction with a GRU head; MOMENT is a T5-style encoder pretrained on 13 time-series domains, processing each channel independently (6$\times$1024\,=\,6144-dim); CrossHAR combines masked reconstruction with contrastive regularization. For zero-shot evaluation, each uses its paper's recommended classifier on training-set embeddings only---the term \emph{zero-shot} means the test dataset is unseen at both training and classifier-fitting stages. These models benefit from group-based scoring but cannot predict unseen label strings.
\emph{Generative language model} (LLaSA) combines a LiMU-BERT encoder with Vicuna-7B and classifies by prompting the LLM with sensor tokens.
These asymmetries are inherent to the architectures; where we deviated from a baseline's protocol, we erred in that baseline's favour (Section~\ref{sec:eval_protocol}). Full adaptation details are in the supplementary material.

\begin{table}[t]
  \centering
  \small
  \setlength{\tabcolsep}{3pt}
  \begin{tabular}{l l r r r}
    \toprule
    Model & Paradigm & Dim & Train. & Infer. \\
    \midrule
    LiMU-BERT~\cite{limubert}  & Masked recon.   & 72   & 72.7K  & 72.7K \\
    MOMENT~\cite{moment}        & General TS      & 6144 & 341.2M & 341.2M \\
    CrossHAR~\cite{crosshar}    & Self-sup HAR    & 72   & 531.4K & 531.4K \\
    LanHAR~\cite{lanhar}        & Text-aligned    & 768  & 13.0M\rlap{$^*$}  & 122.9M \\
    LLaSA~\cite{llasa}          & LLM generative  & ---  & ${\sim}$6.7B & ${\sim}$6.7B \\
    \midrule
    HALO (ours)                  & Text-aligned    & 384  & ${\sim}$35M & ${\sim}$58M \\
    \bottomrule
  \end{tabular}
  \caption{Baseline overview. \emph{Train.}: parameters updated during pretraining and alignment. \emph{Infer.}: Total parameters used at inference, including frozen modules.  HALO's inference total includes a single frozen MiniLM (22.7M) shared for both contrastive alignment and channel conditioning; all text embeddings are precomputable. $^*$LanHAR freezes SciBERT (109.9M) during stage-2 and unfreezes it for fine-tuning. }
  \label{tab:baselines}
  \vspace{-1em}
\end{table}

\textbf{Evaluation protocol.}
\label{sec:eval_protocol}
To handle cross-dataset label inconsistency (e.g., ``jogging'' vs.\ ``running''), we define \emph{synonym groups} mapping variant strings to canonical activities; classifier-based models benefit from group-based scoring, while text-aligned models use exact string match. Novel test activities with no training equivalent form irreducible error floors. We report accuracy and macro F1 for all settings.
\emph{Zero-shot open-set:} predict from the full 87-label training vocabulary (HALO uses per-patch majority voting; classifier-based models use 87-class logits).
\emph{Zero-shot closed-set:} restrict predictions to test-dataset labels.
\emph{1\%/10\% supervised:} fine-tune end-to-end on 1\% or 10\% of labeled test data (balanced subsampling, 80/10/10 split, seed~3431; 20 epochs, AdamW, encoder LR $10^{-5}$, head LR $10^{-3}$, batch~32, patience~5).

\emph{Fairness and input policy.}
Each model receives the richest input its architecture can consume.
HALO evaluates at native rates with rich channel descriptions; LanHAR at 50\,Hz; LiMU-BERT, CrossHAR, MOMENT, and LLaSA use the LIMU-BERT preprocessing pipeline (20\,Hz, 120-step, 6-channel windows); MOMENT additionally left-pads each channel to its expected 512 timesteps per its published protocol.
Conversely, MOMENT benefits from 341M parameters and 6144-dim embeddings, and classifier-based models benefit from group scoring.
To ensure a fair comparison:
(i)~all models receive identical windows, labels, and splits;
(ii)~each baseline uses its recommended classifier;
(iii)~where we deviated from a baseline's original protocol, we chose the option favouring that baseline (e.g., unfreezing CrossHAR, giving MOMENT end-to-end fine-tuning, 10$\times$ LLaSA's per-class budget);
(iv)~classifier-based models use group scoring, while text-aligned models use exact string match.
We report exact match for text-aligned models because open-vocabulary deployment depends on raw label-string retrieval; synonym-group scoring would not reflect real deployment accuracy.
As a direct fairness control, Section~\ref{sec:native_rate_ablation} evaluates HALO under baseline-equivalent inputs (20\,Hz, generic descriptions) and shows it still leads on 3 of 4 settings (Table~\ref{tab:fairness_control}).

\textbf{Implementation details.}
Training proceeds in two stages with a single model jointly across all 10 datasets.
Stage~1 pretrains the IMU encoder (Section~\ref{sec:pretraining}) for 100~epochs (batch~20, AdamW, LR~$10^{-4}$, weight decay~$10^{-5}$, 10 warmup epochs, MAE mask ratio 0.5, contrastive temperature 0.2).
Stage~2 initializes from the pretrained encoder and trains for 200~epochs with 70/15/15 splits per dataset, mixed precision, and gradient accumulation (micro-batch~32, accumulation~16). Key hyperparameters: learned temperature $\alpha_0{=}0.07$, soft-target temperature $\tau_s{=}0.5$, memory bank $Q{=}256$.
Group-balanced sampling mitigates class imbalance (weights inversely proportional to group frequency, capped at $20\times$).
HALO uses a fixed 1.0\,s evaluation patch size for all datasets with no per-dataset tuning; sensitivity is analyzed in Section~\ref{sec:patch_sensitivity}. All training is performed on a single NVIDIA A100 80\,GB GPU; on-device inference profiling uses a smartphone (Section~\ref{sec:scaling}).

\subsection{Overall Performance}

\begin{table}[t]
  \centering
  \small
  \setlength{\tabcolsep}{3pt}
  \begin{tabular}{l cc cc cc cc}
    \toprule
    & \multicolumn{2}{c}{ZS-O}
    & \multicolumn{2}{c}{ZS-C}
    & \multicolumn{2}{c}{1\%}
    & \multicolumn{2}{c}{10\%} \\
    \cmidrule(lr){2-3} \cmidrule(lr){4-5} \cmidrule(lr){6-7} \cmidrule(lr){8-9}
    Model & Acc & F1 & Acc & F1 & Acc & F1 & Acc & F1 \\
    \midrule
    HALO (ours)
      & \textbf{42.0} & \textbf{21.0}
      & \textbf{53.1} & \textbf{37.2}
      & \textbf{76.5} & \textbf{70.0}
      & \textbf{85.7} & \textbf{81.7} \\
    MOMENT
      & 28.3 & 7.2 & 44.7 & 33.2 & 73.8 & 69.8 & 81.3 & 76.8 \\
    CrossHAR
      & 21.2 & 5.5 & 38.2 & 31.7 & 61.1 & 55.2 & 78.6 & 75.2 \\
    LiMU-BERT
      & 21.7 & 7.5 & 33.1 & 23.1 & 35.7 & 22.7 & 51.3 & 43.6 \\
    LanHAR
      & 15.8 & 6.2 & 28.4 & 21.7 & 43.3 & 35.2 & 56.2 & 51.5 \\
    LLaSA$^\dagger$
      & 1.4 & 1.2 & 17.4 & 8.4 & --- & --- & --- & --- \\
    \bottomrule
  \end{tabular}
  \caption{Overall performance in average accuracy (\%) and macro F1 (\%) across 5 main test datasets. \textbf{Bold}: best result for each metric. ZS-O/ZS-C: zero-shot open-set/closed-set settings.
  HALO and LanHAR are evaluated at each dataset's native sampling rate; other baselines use data resampled to 20\,Hz. $^\dagger$LLaSA: published 7B model, zero-shot only (no supervised)
  }
  \label{tab:main_results}
  \vspace{-1em}
\end{table}



\subsubsection{Overall performance in different settings}

Table~\ref{tab:main_results} summarizes average performance. HALO leads on all eight metrics; notably, these gains hold even under baseline-equivalent inputs (20\,Hz, generic descriptions), as confirmed in the fairness control (Table~\ref{tab:fairness_control}). The largest margin is in zero-shot open-set (42.0\% vs.\ MOMENT's 28.3\%, +13.7~pp), where predicting from 87 candidates without test-set knowledge is hardest. Even at 1\% supervision, HALO leads despite MOMENT's 6144-dim embeddings (16$\times$ HALO's 384-dim) giving its classifier substantially more capacity. LanHAR underperforms despite being text-aligned and evaluating at native 50\,Hz---likely because its sensor encoder lacks self-supervised pretraining.

\subsubsection{Results in each dataset}

\begin{table}[t]
  \centering
  \small
  \setlength{\tabcolsep}{3pt}
  \begin{tabular}{l cccc cccc}
    \toprule
    & \multicolumn{4}{c}{Accuracy (\%)} & \multicolumn{4}{c}{Macro F1 (\%)} \\
    \cmidrule(lr){2-5} \cmidrule(lr){6-9}
    Dataset & ZS-O & ZS-C & 1\% & 10\% & ZS-O & ZS-C & 1\% & 10\% \\
    \midrule
    MotionSense & 49.3 & 64.1 & 88.6 & 95.0 & 30.3 & 49.3 & 87.1 & 94.6 \\
    RealWorld & 48.0 & 48.0 & 75.1 & 85.6 & 33.3 & 37.9 & 74.0 & 86.4 \\
    MobiAct & 42.2 & 50.0 & 65.3 & 72.9 & 16.3 & 12.8 & 35.2 & 51.7 \\
    Shoaib & 49.2 & 54.1 & 81.6 & 95.9 & 17.6 & 51.5 & 81.2 & 95.7 \\
    Opportunity & 21.1 & 49.3 & 72.0 & 79.3 & 7.4 & 34.2 & 72.6 & 80.0 \\
    \midrule
    \textbf{Average} & 42.0 & 53.1 & 76.5 & 85.7 & 21.0 & 37.2 & 70.0 & 81.7 \\
    \bottomrule
  \end{tabular}
  \caption{Per-dataset results of HALO on the five main test datasets. ZS-O/ZS-C: zero-shot open/closed-set. All datasets are held out from training.}
  \label{tab:per_dataset}
  \vspace{-1em}
\end{table}

HALO leads zero-shot on all five datasets except Opportunity closed-set, where MOMENT edges ahead (53.9\% vs.\ 49.3\%), possibly because its 6144-dim SVM-RBF classifier benefits from the small label set. MobiAct---the hardest with 13~classes including falls---shows the largest zero-shot gap (HALO 42.2\% vs.\ MOMENT 28.7\% open-set). At 10\% supervision, HALO leads on all 5 datasets; at 1\%, HALO leads on 4 of 5 (MOMENT edges ahead only on Opportunity: 74.3\% vs.\ 72.0\%).

\begin{table}
  \centering
  \small
  \setlength{\tabcolsep}{3pt}
  \begin{tabular}{l cccc cccc}
    \toprule
    & \multicolumn{4}{c}{HARTH} & \multicolumn{4}{c}{VTT-ConIoT} \\
    \cmidrule(lr){2-5} \cmidrule(lr){6-9}
    & ZS-O & ZS-C & 1\% & 10\% & ZS-O & ZS-C & 1\% & 10\% \\
    \midrule
    HALO    & 2.0 & 2.5 & 64.7 & \textbf{78.3} & 1.3 & 2.2 & 1.9 & 16.9 \\
    MOMENT  & 0.4 & 1.9 & 66.7 & 65.4 & 1.6 & 5.2 & \textbf{18.4} & \textbf{34.8} \\
    CrossHAR& 0.3 & 0.9 & 42.4 & 71.9 & 0.7 & 5.0 & 10.6 & 30.9 \\
    LiMU-BERT&0.0 &\textbf{20.2}& 54.4 & 19.1 & 3.4 & \textbf{7.1} & 4.3 & 15.0 \\
    LanHAR  & 0.9 & 1.1 & 3.4 & 70.0 & \textbf{8.3} & 6.9 & 7.7 & 7.7 \\
    LLaSA$^\dagger$& 11.9& 12.5 & --- & --- & --- & 6.9 & --- & --- \\
    \bottomrule
  \end{tabular}
  \caption{Accuracy on severe out-of-domain datasets (\%). ZS-O/ZS-C: zero-shot open/closed-set. All models collapse on HARTH zero-shot (sensor distribution shift); VTT-ConIoT is capped at ${\sim}$50\% by novel activities. $^\dagger$LLaSA: zero-shot only, subsampled.}
  \label{tab:severe_ood}
\end{table}

\subsubsection{Severe out-of-domain transfer performance}
\label{sec:severe_ood}

Table~\ref{tab:severe_ood} reports HARTH and VTT-ConIoT separately. HARTH uses back/thigh-mounted accelerometers only (no gyroscope), with gravity-contaminated data that differs substantially from training; despite 100\% label coverage, this sensor shift causes all models to collapse in zero-shot. VTT-ConIoT is an industrial dataset where only 50\% of activities have training equivalents; the rest (e.g., kneeling work, roll painting) are entirely novel.
On HARTH, all embedding-based models achieve near-zero zero-shot accuracy ($<$3\%), confirming genuine sensor distribution shift rather than a model-specific failure. LLaSA's generative reasoning is less sensitive to signal distribution (11.9\% open-set), but still far from useful. With supervised data, HALO leads at 10\% (78.3\% vs.\ CrossHAR's 71.9\%); at 1\%, MOMENT edges ahead (66.7\% vs.\ 64.7\%).
On VTT-ConIoT, all models score near random on zero-shot ($<$9\%). With supervised data, MOMENT leads (34.8\% at 10\%), benefiting from its general time-series pretraining; HALO's HAR-specific encoder provides strong in-domain transfer but less benefit when the target domain is fundamentally different.

\subsection{Understanding HALO's Performance}

\subsubsection{Ablation study on design components}
\label{sec:component_ablation}

We isolate three architectural and training contributions by disabling each independently from the full model.\footnote{Ablation runs use 100 Stage~2 epochs (vs.\ 200 for the final model) with a single seed; absolute numbers differ slightly from Table~\ref{tab:main_results} but relative deltas are meaningful.} Table~\ref{tab:component_ablation} reports 5-dataset averages.
\emph{Channel-text fusion} (i.e., sensor conditioning, Section~\ref{sec:conditioning}) is the single most important zero-shot component ($-$19.6\,pp ZS-O), as the model must otherwise infer sensor semantics purely from signal patterns; the effect fades with supervision ($-$2.9\,pp at 10\%).
\emph{Text augmentation} has the largest supervised impact ($-$7.0\,pp at 1\%), preventing the text encoder from memorizing exact label strings.
\emph{Signal augmentation} closes a 9.5\,pp generalization gap invisible on in-distribution validation but critical on held-out datasets; on severe OOD the effect strengthens to $-$6.8\,pp at 10\%.

\begin{table}[t]
  \centering
  \footnotesize
  \setlength{\tabcolsep}{3pt}
  \begin{tabular}{l cccc}
    \toprule
    Configuration & ZS-O & ZS-C & 1\% & 10\% \\
    \midrule
    Full model
      & \textbf{44.9} & \textbf{51.9} & \textbf{74.7} & \textbf{85.4} \\
    $-$ Channel-text fusion
      & 25.3 {\footnotesize(--19.6)} & 43.8 {\footnotesize(--8.1)} & 72.9 {\footnotesize(--1.8)} & 82.5 {\footnotesize(--2.9)} \\
    $-$ Signal augmentation
      & 45.3 {\footnotesize(+0.4)} & 53.0 {\footnotesize(+1.1)} & 70.9 {\footnotesize(--3.8)} & 83.5 {\footnotesize(--1.9)} \\
    $-$ Text augmentation
      & 42.7 {\footnotesize(--2.2)} & 49.9 {\footnotesize(--2.0)} & 67.7 {\footnotesize(--7.0)} & 84.2 {\footnotesize(--1.2)} \\
    \bottomrule
  \end{tabular}
  \caption{Ablation study of design components. Each row removes one component while keeping all others fixed. Results report average accuracy (\%) across five main test datasets, with performance deltas relative to the full model shown in parentheses.}
  \label{tab:component_ablation}
\end{table}

\begin{table}
  \centering
  \small
  \setlength{\tabcolsep}{4pt}
  \begin{tabular}{l cccc}
    \toprule
    Configuration & ZS-O & ZS-C & 1\% & 10\% \\
    \midrule
    HALO (native Hz, rich desc.)
      & \textbf{42.0} & \textbf{53.1} & \textbf{76.5} & \textbf{85.7} \\
    HALO (20\,Hz, generic desc.)
      & 30.6 & 49.4 & 71.7 & 84.7 \\
    \midrule
    Best baseline (MOMENT)
      & 28.3 & 44.7 & 73.8 & 81.3 \\
    \bottomrule
  \end{tabular}
  \caption{Analysis of native sampling rate and evaluation fairness with average accuracy (\%) across five main test datasets. In Row 2, HALO is evaluated using the same 20Hz, 6‑channel inputs with generic textual descriptions as four of the five baselines (LanHAR uses 50Hz data). Even under these conservative and standardized inputs, HALO outperforms baselines in three of the four evaluation settings.}
  \label{tab:fairness_control}
\end{table}

\subsubsection{Analysis on native sampling rate and fairness}
\label{sec:native_rate_ablation}

We compare HALO at native rates with rich descriptions against the same model on 20\,Hz resampled data with generic channel names (e.g., ``acc\_x''), matching the LIMU-BERT pipeline used by four of five baselines.
Table~\ref{tab:fairness_control} shows that even under these conservative inputs, HALO leads on 3 of 4 settings (+2.3~pp ZS-O, +4.7~pp ZS-C, +3.4~pp at 10\% over MOMENT); only at 1\% does MOMENT edge ahead.
The native-rate advantage is dataset-dependent: MotionSense and MobiAct show large zero-shot gains (+24.5 and +19.7~pp) where rich channel text aids placement disambiguation, while RealWorld drops ($-$9.9~pp) because its heterogeneous body placements are poorly captured by a single dataset-level description. Supervised performance is less affected (within 3~pp at 10\%).

\begin{figure}
    \centering
    \setlength{\abovecaptionskip}{2pt}
    \setlength{\belowcaptionskip}{-2pt}
  \includegraphics[width=0.8\columnwidth]{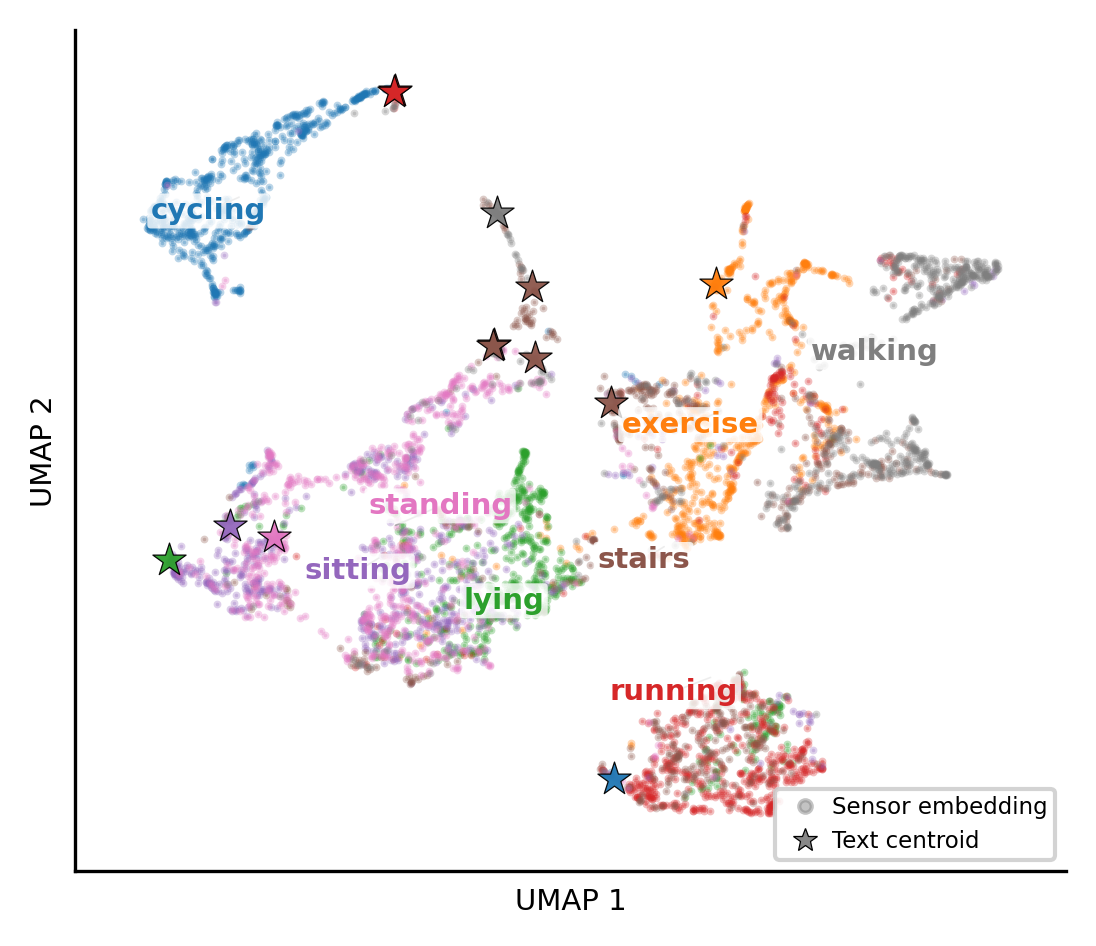}
  \caption{Analysis of embedding space. UMAP projection of learned IMU and text embeddings at epoch~100.
  Circles: per-activity IMU centroids; diamonds: SBERT text centroids; lines connect matched IMU–text pairs.
  }
  \label{fig:embedding_umap}
\end{figure}

\subsubsection{Analysis of embedding space}
\label{sec:embedding_analysis}

Figure~\ref{fig:embedding_umap} visualizes the learned embedding space via UMAP~\cite{umap}. Semantically related groups (e.g., \textit{running}, \textit{stairs}, \textit{walking}) cluster together with tight cross-modal alignment: nearest-neighbor accuracy reaches 76.7\%, positive similarity 0.857, and the similarity gap between matched and unmatched pairs is 0.377. Stationary activities form a distinct cluster separated from locomotion, mirroring kinematic structure. Text centroids align closely with IMU counterparts, confirming the contrastive objective maps both modalities into a shared semantic space.

\subsection{Patch Size Sensitivity}
\label{sec:patch_sensitivity}

\begin{figure}[tb]
    \centering
    \setlength{\abovecaptionskip}{2pt}
    \setlength{\belowcaptionskip}{-4pt}
  \includegraphics[width=0.7\columnwidth]{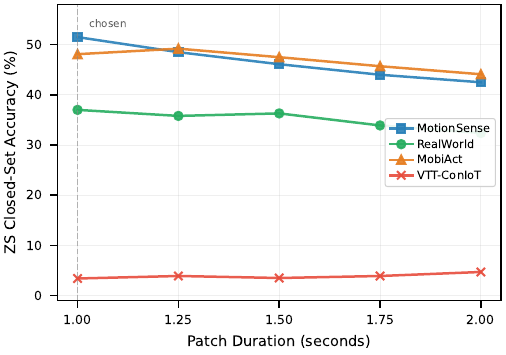}
  \caption{Patch size sensitivity. Zero-shot closed-set accuracy (\%) across candidate patch durations. The dashed vertical line indicates the selected patch size of 1.0\,s. Original datasets are evaluated at 20\,Hz, while Shoaib and Opportunity are evaluated at their native sampling rates.}
  \label{fig:patch_sensitivity}
\end{figure}

HALO uses a fixed 1.0\,s patch size for all results---no per-dataset tuning. Figure~\ref{fig:patch_sensitivity} shows that 1.0--1.25\,s patches consistently perform best; the fixed 1.0\,s choice is within 1.3~pp of the per-dataset optimum across all five datasets. This range aligns with typical gait cycle durations (0.8--1.2\,s), capturing approximately one full motion cycle per patch. Shorter patches (0.75\,s) degrade on complex-activity datasets like Opportunity, likely splitting compound activities across patch boundaries; longer patches (1.5--2.0\,s) show diminishing returns as they average over multiple motion phases. The sensitivity curve is relatively flat between 0.75--1.5\,s, and consistency across sampling rates (20--100\,Hz) confirms that seconds-based patching with adaptive pooling decouples temporal resolution from patch semantics.

\begin{table}
  \centering
  \small
  \setlength{\tabcolsep}{4pt}
  \begin{tabular}{l r cccc}
    \toprule
    Model & Params & ZS-O & ZS-C & 1\% & 10\% \\
    \midrule
    Tiny   & 6M  & 32.8 & 52.1 & 72.6 & \textbf{86.4} \\
    \textbf{Small} & \textbf{35M} & \textbf{46.0} & \textbf{53.4} & \textbf{77.5} & 85.3 \\
    Medium & 63M & 37.8 & 47.7 & 72.6 & 84.5 \\
    \bottomrule
  \end{tabular}
  \caption{Scaling of model size. Average accuracy (\%) on 3 main test datasets at 20\,Hz. The Small model (35M) achieves the best performance, while the Medium model regresses on all zero-shot metrics despite having twice as many parameters.}
  \label{tab:scaling}
\end{table}

\subsection{Scaling of Model Sizes}
\label{sec:scaling}

We train three model sizes---Tiny ($d{=}192$, 4 layers, 6M), Small ($d{=}384$, 8 layers, 35M), and Medium ($d{=}512$, 8 layers, 63M)---with identical training recipes, hyperparameters, and data to isolate the effect of model capacity.
Table~\ref{tab:scaling} shows Small is the sweet spot: Medium regresses by 8.2\,pp on ZS-O despite doubling parameters, achieving the highest \emph{validation} accuracy but the worst held-out generalization---classic overfitting to dataset-specific patterns rather than transferable motion representations. Tiny underfits, falling 13.2\,pp below Small on ZS-O, though it slightly outperforms Small at 10\% supervision (86.4\% vs.\ 85.3\%), suggesting smaller models benefit more from supervised regularization. The ${\sim}$13K-session corpus supports ${\sim}$35M parameters before overfitting dominates; larger HAR corpora could shift this ceiling upward.

\subsection{Inference Overhead on Smartphones}

\begin{table}[t]
\centering
\resizebox{\columnwidth}{!}{
\begin{tabular}{lrcccc}
\toprule
\textbf{Model} & \textbf{Params} & \textbf{Avg (ms)} & \textbf{p95 (ms)} & \textbf{Peak Mem. (MB)} & \textbf{CPU Wt/ms} \\
\midrule
CrossHAR & 110K & 0.29 & 0.30 & 1.20 & 224.0K \\
LiMU-BERT & 110K & 0.31 & 0.36 & 0.27 & 188.0K \\
LLaSA & 105K & 0.33 & 0.46 & 0.02 & 187.2K \\
LanHAR & 8.8M & 0.77 & 0.89 & 0.05 & 71.8K \\
MOMENT & 24.0M & 37.69 & 37.98 & 25.05 & 1.8K \\
\textbf{HALO (ours)} & \textbf{34.0M} & \textbf{7.16} & \textbf{7.37} & \textbf{32.81} & \textbf{2414.4K} \\
\bottomrule
\end{tabular}}
\caption{Inference overhead on an iPhone 16 Pro. All models are benchmarked with same setting: identical 6-channel IMU inputs using Core ML, with 10 warm-up iterations followed by 100 timed iterations. \emph{Params} denotes the profiled model size. \emph{CPU Wt/ms} reports cycle-weighted CPU activity normalized by benchmark duration.}
\label{tab:inference_results}
\vspace{-8pt}
\end{table}

Table~\ref{tab:inference_results} shows that HALO achieves an average inference latency of 7.16\,ms (p95: 7.37\,ms) on an iPhone~16~Pro, well within real-time requirements for on-device activity recognition. Under the same Core~ML deployment setting (identical 6-channel IMU input, 10 warm-up and 100 timed iterations), HALO is $5.26\times$ faster than MOMENT (7.16\,ms vs.\ 37.69\,ms) while jointly modeling all six IMU channels in a single forward pass.
Although HALO is slower than lightweight baselines such as CrossHAR (0.29\,ms) and LiMU-BERT (0.31\,ms) due to its dual-branch Transformer architecture, its latency remains practical for smartphone deployment. The speedup over MOMENT mainly stems from inference structure: MOMENT relies on a large T5-style encoder and requires six separate passes for a 6-channel input, whereas HALO processes all channels jointly. HALO reaches a peak memory usage of 32.81\,MB, and label embeddings are precomputed and cached offline, incurring negligible runtime overhead.


\subsection{Real-World Case Study}
\label{sec:case_study}

\begin{figure}[tbp]
    \centering
    \setlength{\abovecaptionskip}{2pt}
    \setlength{\belowcaptionskip}{-4pt}
    \includegraphics[width=\columnwidth]{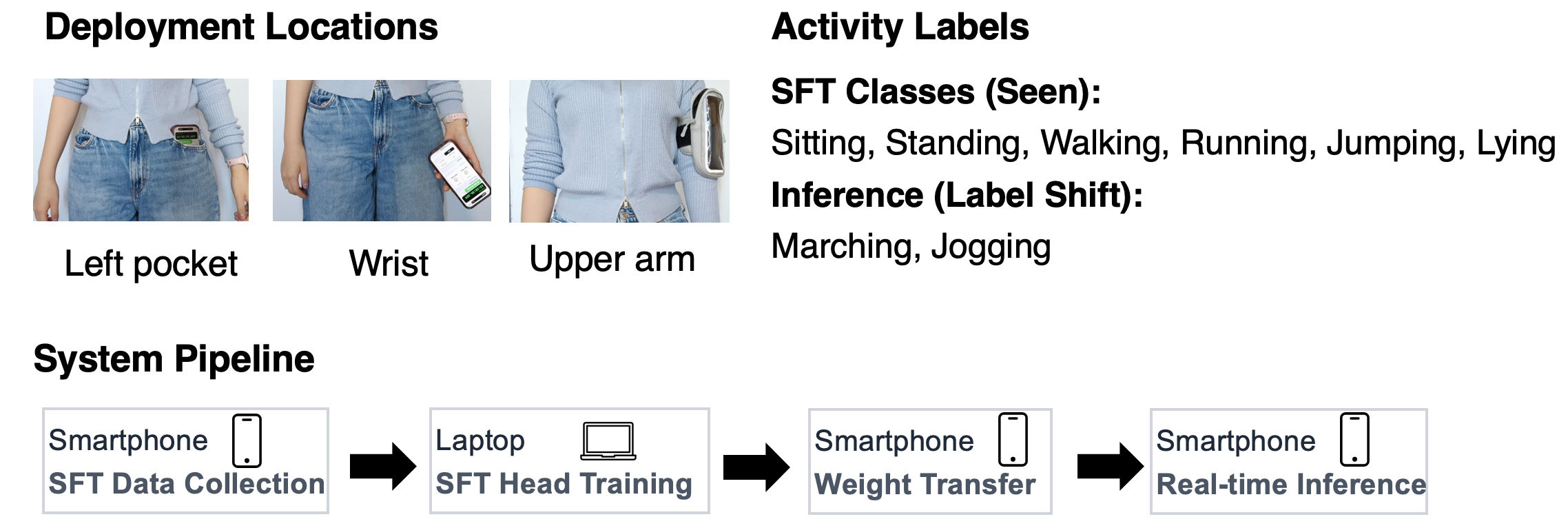}
    \caption{Deployment setting of smartphone-based IMU HAR in the real-world case study.}
    \label{fig:demo_setting}
\end{figure}

We implement the full HALO pipeline as an iOS application and test under continuous placement and label shifts. As shown in Figure~\ref{fig:demo_setting}, the pipeline consists of four stages: (1)~SFT data collection on the smartphone, where the user performs each target activity for 30\,s per class; (2)~SFT head training on a laptop (under 2 minutes for 6 classes); (3)~weight transfer back to the device via a local network; and (4)~real-time inference. The demo covers three body locations (\textit{left pocket}, \textit{wrist}, \textit{upper arm}) with six seen classes (\textit{Sitting}, \textit{Standing}, \textit{Walking}, \textit{Running}, \textit{Jumping}, \textit{Lying}) and two unseen labels (\textit{Marching}, \textit{Jogging}) introduced at inference to test robustness beyond the SFT label space.

\begin{figure}[tbp]
    \centering
    \setlength{\abovecaptionskip}{2pt}
    \setlength{\belowcaptionskip}{-4pt}
    \includegraphics[width=\columnwidth]{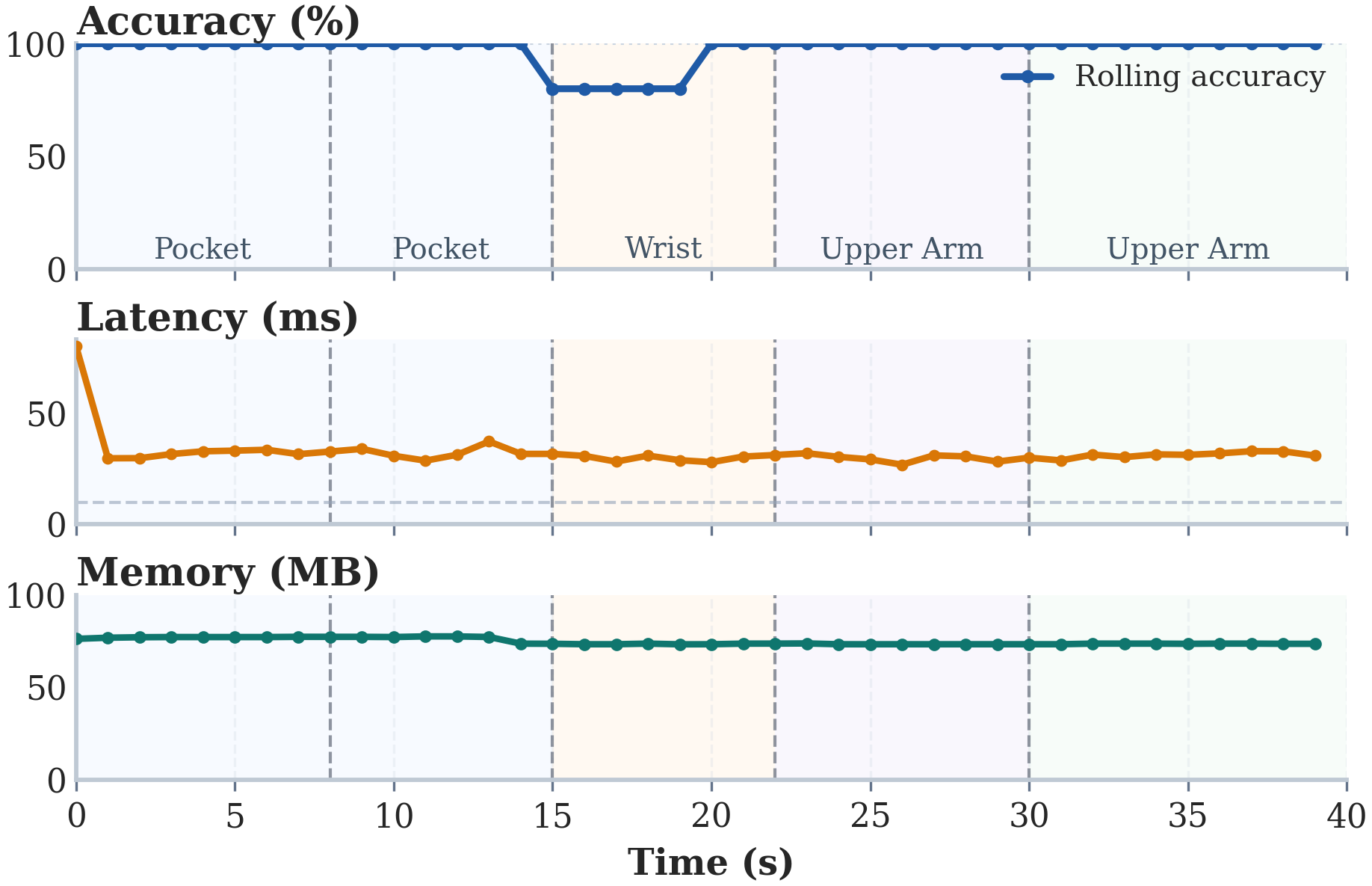}
    \caption{Performance continuous IMU-based HAR on the smartphone.}
    \label{fig:demo_performance}
\end{figure}

Across 40 inference windows (Figure~\ref{fig:demo_performance}), HALO achieves 97.5\% accuracy with only one transient error near the pocket-to-wrist transition, recovering immediately. The placement transition from pocket to wrist represents a significant sensor distribution shift mid-session, yet the model adapts within a single window. The unseen-label phase remains fully correct, confirming that the personalized head preserves semantic generalization even when inference-time labels differ from the SFT classes---the model correctly distinguishes ``Marching'' from ``Walking'' and ``Jogging'' from ``Running'' despite never training on these labels during SFT. Mean end-to-end pipeline latency is 30.35\,ms (P95: 33.66\,ms), higher than the isolated model benchmark (7.16\,ms, Table~\ref{tab:inference_results}) due to app-level overhead; memory stays within 73--78\,MB, and the deployed model is 21.8M parameters (${\sim}$34\,MB on disk)---practical for continuous on-device HAR.

\section{Conclusion and Discussions}
\label{sec:conclusion}


We presented HALO, an IMU foundation model that supports diverse wearable configurations and enables open-vocabulary activity recognition without per-dataset adaptation. With ${\sim}$35M parameters, HALO outperforms five baselines across all eight aggregate metrics.
Our results also motivate several directions for future work. First, moving from dataset-level to per-sample channel–text conditioning could better capture heterogeneous body placements, as evidenced by a 9.9\,pp zero-shot drop on RealWorld with richer global descriptions (Section~\ref{sec:native_rate_ablation}). Second, robustness to severe distribution shift remains an open challenge, particularly for datasets such as HARTH and VTT-ConIoT under zero-shot transfer. Finally, expanding beyond the current 87-label vocabulary and 384‑dimensional embedding space may further enhance semantic coverage and generalization ability of HALO.


\bibliographystyle{ACM-Reference-Format}
\bibliography{reference}


\begin{thebibliography}{39}


\ifx \showCODEN    \undefined \def \showCODEN     #1{\unskip}     \fi
\ifx \showDOI      \undefined \def \showDOI       #1{#1}\fi
\ifx \showISBNx    \undefined \def \showISBNx     #1{\unskip}     \fi
\ifx \showISBNxiii \undefined \def \showISBNxiii  #1{\unskip}     \fi
\ifx \showISSN     \undefined \def \showISSN      #1{\unskip}     \fi
\ifx \showLCCN     \undefined \def \showLCCN      #1{\unskip}     \fi
\ifx \shownote     \undefined \def \shownote      #1{#1}          \fi
\ifx \showarticletitle \undefined \def \showarticletitle #1{#1}   \fi
\ifx \showURL      \undefined \def \showURL       {\relax}        \fi
\providecommand\bibfield[2]{#2}
\providecommand\bibinfo[2]{#2}
\providecommand\natexlab[1]{#1}
\providecommand\showeprint[2][]{arXiv:#2}

\bibitem[\protect\citeauthoryear{Anguita, Ghio, Oneto, Parra, and
  Reyes-Ortiz}{Anguita et~al\mbox{.}}{2013}]%
        {ucihar}
\bibfield{author}{\bibinfo{person}{Davide Anguita}, \bibinfo{person}{Alessandro
  Ghio}, \bibinfo{person}{Luca Oneto}, \bibinfo{person}{Xavier Parra}, {and}
  \bibinfo{person}{Jorge~Luis Reyes-Ortiz}.} \bibinfo{year}{2013}\natexlab{}.
\newblock \showarticletitle{A Public Domain Dataset for Human Activity
  Recognition Using Smartphones}. In \bibinfo{booktitle}{{\em Proceedings of
  the European Symposium on Artificial Neural Networks, Computational
  Intelligence and Machine Learning (ESANN)}}. \bibinfo{address}{Bruges,
  Belgium}.
\newblock


\bibitem[\protect\citeauthoryear{Banos, Garcia, Holgado-Terriza, Damas,
  Pomares, Rojas, Saez, and Villalonga}{Banos et~al\mbox{.}}{2014}]%
        {mhealth}
\bibfield{author}{\bibinfo{person}{Oresti Banos}, \bibinfo{person}{Rafael
  Garcia}, \bibinfo{person}{Juan~A. Holgado-Terriza}, \bibinfo{person}{Miguel
  Damas}, \bibinfo{person}{Hector Pomares}, \bibinfo{person}{Ignacio Rojas},
  \bibinfo{person}{Alejandro Saez}, {and} \bibinfo{person}{Claudia
  Villalonga}.} \bibinfo{year}{2014}\natexlab{}.
\newblock \showarticletitle{{mHealthDroid}: A Novel Framework for Agile
  Development of Mobile Health Applications}. In \bibinfo{booktitle}{{\em
  Proceedings of the 6th International Work-conference on Ambient Assisted
  Living and Daily Activities (IWAAL)}}. \bibinfo{publisher}{Springer},
  \bibinfo{address}{Belfast, UK}, \bibinfo{pages}{91--98}.
\newblock


\bibitem[\protect\citeauthoryear{Barshan and Y{\"u}ksek}{Barshan and
  Y{\"u}ksek}{2014}]%
        {dsads}
\bibfield{author}{\bibinfo{person}{Billur Barshan} {and}
  \bibinfo{person}{Murat~C. Y{\"u}ksek}.} \bibinfo{year}{2014}\natexlab{}.
\newblock \showarticletitle{Recognizing Daily and Sports Activities in Two
  Open-Source Machine Learning Environments Using Body-Worn Sensor Units}.
\newblock \bibinfo{journal}{{\it Comput. J.}} \bibinfo{volume}{57},
  \bibinfo{number}{11} (\bibinfo{year}{2014}), \bibinfo{pages}{1649--1667}.
\newblock


\bibitem[\protect\citeauthoryear{Chang, Di~Campli San~Vito, Musolesi, and
  Mascolo}{Chang et~al\mbox{.}}{2024}]%
        {crosshar}
\bibfield{author}{\bibinfo{person}{Zhiqiang Chang}, \bibinfo{person}{Patrizia
  Di~Campli San~Vito}, \bibinfo{person}{Mirco Musolesi}, {and}
  \bibinfo{person}{Cecilia Mascolo}.} \bibinfo{year}{2024}\natexlab{}.
\newblock \showarticletitle{{CrossHAR}: Generalizing Cross-dataset Human
  Activity Recognition via Hierarchical Self-Supervised Pretraining}.
\newblock \bibinfo{journal}{{\em Proceedings of the ACM on Interactive, Mobile,
  Wearable and Ubiquitous Technologies\/}} \bibinfo{volume}{8},
  \bibinfo{number}{1} (\bibinfo{year}{2024}), \bibinfo{pages}{1--27}.
\newblock


\bibitem[\protect\citeauthoryear{Chen, Kornblith, Norouzi, and Hinton}{Chen
  et~al\mbox{.}}{2020}]%
        {simclr}
\bibfield{author}{\bibinfo{person}{Ting Chen}, \bibinfo{person}{Simon
  Kornblith}, \bibinfo{person}{Mohammad Norouzi}, {and}
  \bibinfo{person}{Geoffrey Hinton}.} \bibinfo{year}{2020}\natexlab{}.
\newblock \showarticletitle{A Simple Framework for Contrastive Learning of
  Visual Representations}. In \bibinfo{booktitle}{{\em Proceedings of the 37th
  International Conference on Machine Learning (ICML)}}.
  \bibinfo{publisher}{PMLR}, \bibinfo{address}{Virtual},
  \bibinfo{pages}{1597--1607}.
\newblock


\bibitem[\protect\citeauthoryear{Devlin, Chang, Lee, and Toutanova}{Devlin
  et~al\mbox{.}}{2019}]%
        {bert}
\bibfield{author}{\bibinfo{person}{Jacob Devlin}, \bibinfo{person}{Ming-Wei
  Chang}, \bibinfo{person}{Kenton Lee}, {and} \bibinfo{person}{Kristina
  Toutanova}.} \bibinfo{year}{2019}\natexlab{}.
\newblock \showarticletitle{{BERT}: Pre-training of Deep Bidirectional
  Transformers for Language Understanding}. In \bibinfo{booktitle}{{\em
  Proceedings of the 2019 Conference of the North American Chapter of the
  Association for Computational Linguistics (NAACL)}}.
  \bibinfo{publisher}{Association for Computational Linguistics},
  \bibinfo{address}{Minneapolis, MN}, \bibinfo{pages}{4171--4186}.
\newblock


\bibitem[\protect\citeauthoryear{Girdhar, El-Nouby, Liu, Singh, Alwala, Joulin,
  and Misra}{Girdhar et~al\mbox{.}}{2023}]%
        {imagebind}
\bibfield{author}{\bibinfo{person}{Rohit Girdhar}, \bibinfo{person}{Alaaeldin
  El-Nouby}, \bibinfo{person}{Zhuang Liu}, \bibinfo{person}{Mannat Singh},
  \bibinfo{person}{Kalyan~Vasudev Alwala}, \bibinfo{person}{Armand Joulin},
  {and} \bibinfo{person}{Ishan Misra}.} \bibinfo{year}{2023}\natexlab{}.
\newblock \showarticletitle{{ImageBind}: One Embedding Space to Bind Them All}.
  In \bibinfo{booktitle}{{\em Proceedings of the IEEE/CVF Conference on
  Computer Vision and Pattern Recognition (CVPR)}}. \bibinfo{publisher}{IEEE},
  \bibinfo{address}{Vancouver, Canada}, \bibinfo{pages}{15180--15190}.
\newblock


\bibitem[\protect\citeauthoryear{Goswami, Szafer, Choudhry, Cai, Li, and
  Dubrawski}{Goswami et~al\mbox{.}}{2024}]%
        {moment}
\bibfield{author}{\bibinfo{person}{Mononito Goswami}, \bibinfo{person}{Konrad
  Szafer}, \bibinfo{person}{Arjun Choudhry}, \bibinfo{person}{Yifu Cai},
  \bibinfo{person}{Shuo Li}, {and} \bibinfo{person}{Artur Dubrawski}.}
  \bibinfo{year}{2024}\natexlab{}.
\newblock \showarticletitle{{MOMENT}: A Family of Open Time-Series Foundation
  Models}. In \bibinfo{booktitle}{{\em Proceedings of the 41st International
  Conference on Machine Learning (ICML)}}. \bibinfo{publisher}{PMLR},
  \bibinfo{address}{Vienna, Austria}, \bibinfo{pages}{16115--16152}.
\newblock


\bibitem[\protect\citeauthoryear{He, Chen, Xie, Li, Doll{\'a}r, and
  Girshick}{He et~al\mbox{.}}{2022}]%
        {mae}
\bibfield{author}{\bibinfo{person}{Kaiming He}, \bibinfo{person}{Xinlei Chen},
  \bibinfo{person}{Saining Xie}, \bibinfo{person}{Yanghao Li},
  \bibinfo{person}{Piotr Doll{\'a}r}, {and} \bibinfo{person}{Ross Girshick}.}
  \bibinfo{year}{2022}\natexlab{}.
\newblock \showarticletitle{Masked Autoencoders Are Scalable Vision Learners}.
  In \bibinfo{booktitle}{{\em Proceedings of the IEEE/CVF Conference on
  Computer Vision and Pattern Recognition (CVPR)}}. \bibinfo{publisher}{IEEE},
  \bibinfo{address}{New Orleans, LA}, \bibinfo{pages}{16000--16009}.
\newblock


\bibitem[\protect\citeauthoryear{He, Fan, Wu, Xie, and Girshick}{He
  et~al\mbox{.}}{2020}]%
        {moco}
\bibfield{author}{\bibinfo{person}{Kaiming He}, \bibinfo{person}{Haoqi Fan},
  \bibinfo{person}{Yuxin Wu}, \bibinfo{person}{Saining Xie}, {and}
  \bibinfo{person}{Ross Girshick}.} \bibinfo{year}{2020}\natexlab{}.
\newblock \showarticletitle{Momentum Contrast for Unsupervised Visual
  Representation Learning}. In \bibinfo{booktitle}{{\em Proceedings of the
  IEEE/CVF Conference on Computer Vision and Pattern Recognition (CVPR)}}.
  \bibinfo{publisher}{IEEE}, \bibinfo{address}{Seattle, WA},
  \bibinfo{pages}{9729--9738}.
\newblock


\bibitem[\protect\citeauthoryear{Imran, Khan, Biswas, and Islam}{Imran
  et~al\mbox{.}}{2025}]%
        {llasa}
\bibfield{author}{\bibinfo{person}{Sheikh~Asif Imran},
  \bibinfo{person}{Mohammad Nur~Hossain Khan}, \bibinfo{person}{Subrata
  Biswas}, {and} \bibinfo{person}{Bashima Islam}.}
  \bibinfo{year}{2025}\natexlab{}.
\newblock \showarticletitle{{LLaSA}: A Sensor-Aware {LLM} for Natural Language
  Reasoning of Human Activity from {IMU} Data}. In \bibinfo{booktitle}{{\em
  Companion of the 2025 ACM International Joint Conference on Pervasive and
  Ubiquitous Computing (UbiComp)}}. \bibinfo{publisher}{ACM}.
\newblock


\bibitem[\protect\citeauthoryear{Kwapisz, Weiss, and Moore}{Kwapisz
  et~al\mbox{.}}{2011}]%
        {wisdm}
\bibfield{author}{\bibinfo{person}{Jennifer~R. Kwapisz},
  \bibinfo{person}{Gary~M. Weiss}, {and} \bibinfo{person}{Samuel~A. Moore}.}
  \bibinfo{year}{2011}\natexlab{}.
\newblock \showarticletitle{Activity Recognition Using Cell Phone
  Accelerometers}.
\newblock \bibinfo{journal}{{\em ACM SIGKDD Explorations Newsletter\/}}
  \bibinfo{volume}{12}, \bibinfo{number}{2} (\bibinfo{year}{2011}),
  \bibinfo{pages}{74--82}.
\newblock


\bibitem[\protect\citeauthoryear{Li, Li, Hu, Guo, and Yu}{Li
  et~al\mbox{.}}{2024}]%
        {lanhar}
\bibfield{author}{\bibinfo{person}{Quan Li}, \bibinfo{person}{Kecheng Li},
  \bibinfo{person}{Yuqian Hu}, \bibinfo{person}{Bin Guo}, {and}
  \bibinfo{person}{Zhiwen Yu}.} \bibinfo{year}{2024}\natexlab{}.
\newblock \showarticletitle{{LanHAR}: Language-Based Zero-Shot Human Activity
  Recognition via Large Language Models}.
\newblock \bibinfo{journal}{{\em Proceedings of the ACM on Interactive, Mobile,
  Wearable and Ubiquitous Technologies\/}} \bibinfo{volume}{8},
  \bibinfo{number}{4} (\bibinfo{year}{2024}), \bibinfo{pages}{1--25}.
\newblock


\bibitem[\protect\citeauthoryear{Logacjov, Bach, Kongsvold, B{\aa}rdstu, and
  Mork}{Logacjov et~al\mbox{.}}{2021}]%
        {harth}
\bibfield{author}{\bibinfo{person}{Aleksej Logacjov}, \bibinfo{person}{Kerstin
  Bach}, \bibinfo{person}{Atle Kongsvold}, \bibinfo{person}{Hilde~Bremseth
  B{\aa}rdstu}, {and} \bibinfo{person}{Paul~Jarle Mork}.}
  \bibinfo{year}{2021}\natexlab{}.
\newblock \showarticletitle{{HARTH}: A Human Activity Recognition Dataset for
  Machine Learning}.
\newblock \bibinfo{journal}{{\em Sensors\/}} \bibinfo{volume}{21},
  \bibinfo{number}{21} (\bibinfo{year}{2021}), \bibinfo{pages}{7261}.
\newblock


\bibitem[\protect\citeauthoryear{M{\"a}kel{\"a}, L{\"a}ms{\"a}, Ker{\"a}nen,
  Liikka, Ronkainen, Peltola, H{\"a}iki{\"o}, J{\"a}rvinen, and
  L{\'o}pez}{M{\"a}kel{\"a} et~al\mbox{.}}{2021}]%
        {vttconiot}
\bibfield{author}{\bibinfo{person}{Satu-Marja M{\"a}kel{\"a}},
  \bibinfo{person}{Ari L{\"a}ms{\"a}}, \bibinfo{person}{Joonas~S. Ker{\"a}nen},
  \bibinfo{person}{Jari Liikka}, \bibinfo{person}{Jari Ronkainen},
  \bibinfo{person}{Jari Peltola}, \bibinfo{person}{Jukka H{\"a}iki{\"o}},
  \bibinfo{person}{Sami J{\"a}rvinen}, {and} \bibinfo{person}{Miguel~Bordallo
  L{\'o}pez}.} \bibinfo{year}{2021}\natexlab{}.
\newblock \showarticletitle{Introducing {VTT-ConIoT}: A Realistic Dataset for
  Activity Recognition of Construction Workers Using {IMU} Devices}.
\newblock \bibinfo{journal}{{\em Sustainability\/}} \bibinfo{volume}{14},
  \bibinfo{number}{1} (\bibinfo{year}{2021}), \bibinfo{pages}{220}.
\newblock
\showDOI{%
\url{https://doi.org/10.3390/su14010220}}


\bibitem[\protect\citeauthoryear{Malekzadeh, Clegg, Cavallaro, and
  Haddadi}{Malekzadeh et~al\mbox{.}}{2019}]%
        {motionsense}
\bibfield{author}{\bibinfo{person}{Mohammad Malekzadeh},
  \bibinfo{person}{Richard~G. Clegg}, \bibinfo{person}{Andrea Cavallaro}, {and}
  \bibinfo{person}{Hamed Haddadi}.} \bibinfo{year}{2019}\natexlab{}.
\newblock \showarticletitle{Mobile Sensor Data Anonymization}. In
  \bibinfo{booktitle}{{\em Proceedings of the International Conference on
  Internet of Things Design and Implementation (IoTDI)}}.
  \bibinfo{publisher}{ACM}, \bibinfo{address}{Montreal, Canada},
  \bibinfo{pages}{49--58}.
\newblock


\bibitem[\protect\citeauthoryear{McInnes, Healy, and Melville}{McInnes
  et~al\mbox{.}}{2018}]%
        {umap}
\bibfield{author}{\bibinfo{person}{Leland McInnes}, \bibinfo{person}{John
  Healy}, {and} \bibinfo{person}{James Melville}.}
  \bibinfo{year}{2018}\natexlab{}.
\newblock \showarticletitle{{UMAP}: Uniform Manifold Approximation and
  Projection for Dimension Reduction}.
\newblock \bibinfo{journal}{{\em arXiv preprint arXiv:1802.03426\/}}
  (\bibinfo{year}{2018}).
\newblock


\bibitem[\protect\citeauthoryear{Micucci, Mobilio, and Napoletano}{Micucci
  et~al\mbox{.}}{2017}]%
        {unimibi}
\bibfield{author}{\bibinfo{person}{Daniela Micucci}, \bibinfo{person}{Marco
  Mobilio}, {and} \bibinfo{person}{Paolo Napoletano}.}
  \bibinfo{year}{2017}\natexlab{}.
\newblock \showarticletitle{{UniMiB SHAR}: A Dataset for Human Activity
  Recognition Using Acceleration Data from Smartphones}.
\newblock \bibinfo{journal}{{\em Applied Sciences\/}} \bibinfo{volume}{7},
  \bibinfo{number}{10} (\bibinfo{year}{2017}), \bibinfo{pages}{1101}.
\newblock


\bibitem[\protect\citeauthoryear{Nie, Nguyen, Sinthong, and Kalagnanam}{Nie
  et~al\mbox{.}}{2023}]%
        {patchtst}
\bibfield{author}{\bibinfo{person}{Yuqi Nie}, \bibinfo{person}{Nam~H. Nguyen},
  \bibinfo{person}{Phanwadee Sinthong}, {and} \bibinfo{person}{Jayant
  Kalagnanam}.} \bibinfo{year}{2023}\natexlab{}.
\newblock \showarticletitle{A Time Series is Worth 64 Words: Long-term
  Forecasting with Transformers}. In \bibinfo{booktitle}{{\em The Eleventh
  International Conference on Learning Representations (ICLR)}}.
  \bibinfo{publisher}{OpenReview.net}, \bibinfo{address}{Kigali, Rwanda},
  \bibinfo{pages}{1--22}.
\newblock


\bibitem[\protect\citeauthoryear{Ord{\'o}{\~n}ez and Roggen}{Ord{\'o}{\~n}ez
  and Roggen}{2016}]%
        {deepconvlstm}
\bibfield{author}{\bibinfo{person}{Francisco~Javier Ord{\'o}{\~n}ez} {and}
  \bibinfo{person}{Daniel Roggen}.} \bibinfo{year}{2016}\natexlab{}.
\newblock \showarticletitle{Deep Convolutional and {LSTM} Recurrent Neural
  Networks for Multimodal Wearable Activity Recognition}.
\newblock \bibinfo{journal}{{\em Sensors\/}} \bibinfo{volume}{16},
  \bibinfo{number}{1} (\bibinfo{year}{2016}), \bibinfo{pages}{115}.
\newblock


\bibitem[\protect\citeauthoryear{Ouyang, Shuai, Zhou, Shi, Xie, Xing, and
  Huang}{Ouyang et~al\mbox{.}}{2022}]%
        {cosmo}
\bibfield{author}{\bibinfo{person}{Xiaomin Ouyang}, \bibinfo{person}{Xian
  Shuai}, \bibinfo{person}{Jiayu Zhou}, \bibinfo{person}{Ivy~Wang Shi},
  \bibinfo{person}{Zhiyuan Xie}, \bibinfo{person}{Guoliang Xing}, {and}
  \bibinfo{person}{Jianwei Huang}.} \bibinfo{year}{2022}\natexlab{}.
\newblock \showarticletitle{Cosmo: Contrastive Fusion Learning with Small Data
  for Multimodal Human Activity Recognition}. In \bibinfo{booktitle}{{\em
  Proceedings of the 28th Annual International Conference on Mobile Computing
  and Networking (MobiCom)}}. \bibinfo{publisher}{ACM},
  \bibinfo{pages}{324--337}.
\newblock


\bibitem[\protect\citeauthoryear{Ouyang, Xie, Zhou, Huang, and Xing}{Ouyang
  et~al\mbox{.}}{2021}]%
        {clusterfl}
\bibfield{author}{\bibinfo{person}{Xiaomin Ouyang}, \bibinfo{person}{Zhiyuan
  Xie}, \bibinfo{person}{Jiayu Zhou}, \bibinfo{person}{Jianwei Huang}, {and}
  \bibinfo{person}{Guoliang Xing}.} \bibinfo{year}{2021}\natexlab{}.
\newblock \showarticletitle{{ClusterFL}: A Similarity-Aware Federated Learning
  System for Human Activity Recognition}. In \bibinfo{booktitle}{{\em
  Proceedings of the 19th Annual International Conference on Mobile Systems,
  Applications, and Services (ACM MobiSys)}}. \bibinfo{publisher}{ACM},
  \bibinfo{pages}{54--67}.
\newblock
\showDOI{%
\url{https://doi.org/10.1145/3458864.3467681}}


\bibitem[\protect\citeauthoryear{Radford, Kim, Hallacy, Ramesh, Goh, Agarwal,
  Sastry, Askell, Mishkin, Clark, Krueger, and Sutskever}{Radford
  et~al\mbox{.}}{2021}]%
        {clip}
\bibfield{author}{\bibinfo{person}{Alec Radford}, \bibinfo{person}{Jong~Wook
  Kim}, \bibinfo{person}{Chris Hallacy}, \bibinfo{person}{Aditya Ramesh},
  \bibinfo{person}{Gabriel Goh}, \bibinfo{person}{Sandhini Agarwal},
  \bibinfo{person}{Girish Sastry}, \bibinfo{person}{Amanda Askell},
  \bibinfo{person}{Pamela Mishkin}, \bibinfo{person}{Jack Clark},
  \bibinfo{person}{Gretchen Krueger}, {and} \bibinfo{person}{Ilya Sutskever}.}
  \bibinfo{year}{2021}\natexlab{}.
\newblock \showarticletitle{Learning Transferable Visual Models From Natural
  Language Supervision}. In \bibinfo{booktitle}{{\em Proceedings of the 38th
  International Conference on Machine Learning (ICML)}}.
  \bibinfo{publisher}{PMLR}, \bibinfo{address}{Virtual},
  \bibinfo{pages}{8748--8763}.
\newblock


\bibitem[\protect\citeauthoryear{Reimers and Gurevych}{Reimers and
  Gurevych}{2019}]%
        {sbert}
\bibfield{author}{\bibinfo{person}{Nils Reimers} {and} \bibinfo{person}{Iryna
  Gurevych}.} \bibinfo{year}{2019}\natexlab{}.
\newblock \showarticletitle{Sentence-{BERT}: Sentence Embeddings using Siamese
  {BERT}-Networks}. In \bibinfo{booktitle}{{\em Proceedings of the 2019
  Conference on Empirical Methods in Natural Language Processing (EMNLP)}}.
  \bibinfo{publisher}{Association for Computational Linguistics},
  \bibinfo{address}{Hong Kong, China}, \bibinfo{pages}{3982--3992}.
\newblock


\bibitem[\protect\citeauthoryear{Reiss and Stricker}{Reiss and
  Stricker}{2012}]%
        {pamap2}
\bibfield{author}{\bibinfo{person}{Attila Reiss} {and} \bibinfo{person}{Didier
  Stricker}.} \bibinfo{year}{2012}\natexlab{}.
\newblock \showarticletitle{Introducing a New Benchmarked Dataset for Activity
  Monitoring}. In \bibinfo{booktitle}{{\em Proceedings of the 16th
  International Symposium on Wearable Computers (ISWC)}}.
  \bibinfo{publisher}{IEEE}, \bibinfo{address}{Newcastle, UK},
  \bibinfo{pages}{108--109}.
\newblock


\bibitem[\protect\citeauthoryear{Reyes-Ortiz, Oneto, Sam{\`a}, Parra, and
  Anguita}{Reyes-Ortiz et~al\mbox{.}}{2016}]%
        {hapt}
\bibfield{author}{\bibinfo{person}{Jorge-Luis Reyes-Ortiz},
  \bibinfo{person}{Luca Oneto}, \bibinfo{person}{Albert Sam{\`a}},
  \bibinfo{person}{Xavier Parra}, {and} \bibinfo{person}{Davide Anguita}.}
  \bibinfo{year}{2016}\natexlab{}.
\newblock \showarticletitle{Transition-Aware Human Activity Recognition Using
  Smartphones}.
\newblock \bibinfo{journal}{{\em Neurocomputing\/}}  \bibinfo{volume}{171}
  (\bibinfo{year}{2016}), \bibinfo{pages}{754--767}.
\newblock


\bibitem[\protect\citeauthoryear{Roggen, Calatroni, Rossi, Holleczek,
  F{\"o}rster, Tr{\"o}ster, Lukowicz, Bannach, Pirkl, Ferscha,
  et~al\mbox{.}}{Roggen et~al\mbox{.}}{2010}]%
        {opportunity}
\bibfield{author}{\bibinfo{person}{Daniel Roggen}, \bibinfo{person}{Alberto
  Calatroni}, \bibinfo{person}{Mirco Rossi}, \bibinfo{person}{Thomas
  Holleczek}, \bibinfo{person}{Kilian F{\"o}rster}, \bibinfo{person}{Gerhard
  Tr{\"o}ster}, \bibinfo{person}{Paul Lukowicz}, \bibinfo{person}{David
  Bannach}, \bibinfo{person}{Gerald Pirkl}, \bibinfo{person}{Alois Ferscha},
  {et~al\mbox{.}}} \bibinfo{year}{2010}\natexlab{}.
\newblock \showarticletitle{Collecting Complex Activity Datasets in Highly Rich
  Networked Sensor Environments}. In \bibinfo{booktitle}{{\em Proceedings of
  the 7th International Conference on Networked Sensing Systems (INSS)}}.
  \bibinfo{publisher}{IEEE}, \bibinfo{pages}{233--240}.
\newblock


\bibitem[\protect\citeauthoryear{Shoaib, Bosch, Incel, Scholten, and
  Havinga}{Shoaib et~al\mbox{.}}{2014}]%
        {shoaib}
\bibfield{author}{\bibinfo{person}{Muhammad Shoaib}, \bibinfo{person}{Stephan
  Bosch}, \bibinfo{person}{Ozlem~Durmaz Incel}, \bibinfo{person}{Hans
  Scholten}, {and} \bibinfo{person}{Paul J.~M. Havinga}.}
  \bibinfo{year}{2014}\natexlab{}.
\newblock \showarticletitle{Fusion of Smartphone Motion Sensors for Physical
  Activity Recognition}.
\newblock \bibinfo{journal}{{\em Sensors\/}} \bibinfo{volume}{14},
  \bibinfo{number}{6} (\bibinfo{year}{2014}), \bibinfo{pages}{10146--10176}.
\newblock


\bibitem[\protect\citeauthoryear{Sikder, Chowdhury, and Nahid}{Sikder
  et~al\mbox{.}}{2021}]%
        {kuhar}
\bibfield{author}{\bibinfo{person}{Nilufar Sikder}, \bibinfo{person}{Muhammad
  Sakib~Ibne Chowdhury}, {and} \bibinfo{person}{Abdullah-Al Nahid}.}
  \bibinfo{year}{2021}\natexlab{}.
\newblock \showarticletitle{{KU-HAR}: An Open Dataset for Heterogeneous Human
  Activity Recognition}.
\newblock \bibinfo{journal}{{\em Pattern Recognition Letters\/}}
  \bibinfo{volume}{146} (\bibinfo{year}{2021}), \bibinfo{pages}{46--54}.
\newblock


\bibitem[\protect\citeauthoryear{Stisen, Blunck, Bhattacharya, Prentow,
  Kj{\ae}rgaard, Dey, Sonne, and Jensen}{Stisen et~al\mbox{.}}{2015}]%
        {hhar}
\bibfield{author}{\bibinfo{person}{Allan Stisen}, \bibinfo{person}{Henrik
  Blunck}, \bibinfo{person}{Sourav Bhattacharya}, \bibinfo{person}{Thor~Siiger
  Prentow}, \bibinfo{person}{Mikkel~Bent Kj{\ae}rgaard}, \bibinfo{person}{Anind
  Dey}, \bibinfo{person}{Tobias Sonne}, {and} \bibinfo{person}{Mads~M{\o}ller
  Jensen}.} \bibinfo{year}{2015}\natexlab{}.
\newblock \showarticletitle{Smart Devices Are Different: Assessing and
  Mitigating Mobile Sensing Heterogeneities for Activity Recognition}. In
  \bibinfo{booktitle}{{\em Proceedings of the 13th ACM Conference on Embedded
  Networked Sensor Systems (SenSys)}}. \bibinfo{publisher}{ACM},
  \bibinfo{address}{Seoul, South Korea}, \bibinfo{pages}{127--140}.
\newblock


\bibitem[\protect\citeauthoryear{Sztyler and Stuckenschmidt}{Sztyler and
  Stuckenschmidt}{2016}]%
        {realworld}
\bibfield{author}{\bibinfo{person}{Timo Sztyler} {and} \bibinfo{person}{Heiner
  Stuckenschmidt}.} \bibinfo{year}{2016}\natexlab{}.
\newblock \showarticletitle{On-Body Localization of Wearable Devices: An
  Investigation of Position-Aware Activity Recognition}. In
  \bibinfo{booktitle}{{\em Proceedings of the IEEE International Conference on
  Pervasive Computing and Communications (PerCom)}}. \bibinfo{publisher}{IEEE},
  \bibinfo{address}{Sydney, Australia}, \bibinfo{pages}{1--9}.
\newblock


\bibitem[\protect\citeauthoryear{van~den Oord, Li, and Vinyals}{van~den Oord
  et~al\mbox{.}}{2018}]%
        {infonce}
\bibfield{author}{\bibinfo{person}{Aaron van~den Oord}, \bibinfo{person}{Yazhe
  Li}, {and} \bibinfo{person}{Oriol Vinyals}.} \bibinfo{year}{2018}\natexlab{}.
\newblock \showarticletitle{Representation Learning with Contrastive Predictive
  Coding}.
\newblock \bibinfo{journal}{{\em arXiv preprint arXiv:1807.03748\/}}
  (\bibinfo{year}{2018}).
\newblock


\bibitem[\protect\citeauthoryear{Vaswani, Shazeer, Parmar, Uszkoreit, Jones,
  Gomez, Kaiser, and Polosukhin}{Vaswani et~al\mbox{.}}{2017}]%
        {transformer}
\bibfield{author}{\bibinfo{person}{Ashish Vaswani}, \bibinfo{person}{Noam
  Shazeer}, \bibinfo{person}{Niki Parmar}, \bibinfo{person}{Jakob Uszkoreit},
  \bibinfo{person}{Llion Jones}, \bibinfo{person}{Aidan~N. Gomez},
  \bibinfo{person}{{\L}ukasz Kaiser}, {and} \bibinfo{person}{Illia
  Polosukhin}.} \bibinfo{year}{2017}\natexlab{}.
\newblock \showarticletitle{Attention Is All You Need}. In
  \bibinfo{booktitle}{{\em Advances in Neural Information Processing Systems 30
  (NeurIPS)}}. \bibinfo{publisher}{Curran Associates}, \bibinfo{address}{Long
  Beach, CA}, \bibinfo{pages}{5998--6008}.
\newblock


\bibitem[\protect\citeauthoryear{Vavoulas, Chatzaki, Malliotakis, Pediaditis,
  and Tsiknakis}{Vavoulas et~al\mbox{.}}{2016}]%
        {mobiact}
\bibfield{author}{\bibinfo{person}{George Vavoulas},
  \bibinfo{person}{Charikleia Chatzaki}, \bibinfo{person}{Thodoris
  Malliotakis}, \bibinfo{person}{Matthew Pediaditis}, {and}
  \bibinfo{person}{Manolis Tsiknakis}.} \bibinfo{year}{2016}\natexlab{}.
\newblock \showarticletitle{The {MobiAct} Dataset: Recognition of Activities of
  Daily Living Using Smartphones}. In \bibinfo{booktitle}{{\em Proceedings of
  the International Conference on Information and Communication Technologies
  for Ageing Well and e-Health (ICT4AWE)}}. \bibinfo{publisher}{SciTePress},
  \bibinfo{address}{Rome, Italy}, \bibinfo{pages}{143--151}.
\newblock


\bibitem[\protect\citeauthoryear{Xu, Zhou, Tan, and Li}{Xu
  et~al\mbox{.}}{2023}]%
        {unihar}
\bibfield{author}{\bibinfo{person}{Huatao Xu}, \bibinfo{person}{Pengfei Zhou},
  \bibinfo{person}{Rui Tan}, {and} \bibinfo{person}{Mo Li}.}
  \bibinfo{year}{2023}\natexlab{}.
\newblock \showarticletitle{Practically Adopting Human Activity Recognition}.
  In \bibinfo{booktitle}{{\em Proceedings of the 29th Annual International
  Conference on Mobile Computing and Networking (ACM MobiCom)}}.
  \bibinfo{publisher}{ACM}, \bibinfo{pages}{10--21}.
\newblock
\showDOI{%
\url{https://doi.org/10.1145/3570361.3613299}}


\bibitem[\protect\citeauthoryear{Xu, Zhou, Tan, Li, and Shen}{Xu
  et~al\mbox{.}}{2021}]%
        {limubert}
\bibfield{author}{\bibinfo{person}{Huatao Xu}, \bibinfo{person}{Pengfei Zhou},
  \bibinfo{person}{Rui Tan}, \bibinfo{person}{Mo Li}, {and}
  \bibinfo{person}{Guobin Shen}.} \bibinfo{year}{2021}\natexlab{}.
\newblock \showarticletitle{{LIMU-BERT}: Unleashing the Potential of Unlabeled
  Data for {IMU} Sensing Applications}. In \bibinfo{booktitle}{{\em Proceedings
  of the 19th ACM Conference on Embedded Networked Sensor Systems (SenSys)}}.
  \bibinfo{publisher}{ACM}, \bibinfo{address}{Coimbra, Portugal},
  \bibinfo{pages}{220--233}.
\newblock


\bibitem[\protect\citeauthoryear{Zhang, Maddix, Wang, Gupta, Mahoney, Goyal,
  Shchur, Merrill, Turkmen, Wen, Rajan, Luo, Faloutsos, and Wang}{Zhang
  et~al\mbox{.}}{2025}]%
        {mitra}
\bibfield{author}{\bibinfo{person}{Xiyuan Zhang}, \bibinfo{person}{Danielle
  Maddix}, \bibinfo{person}{Hao Wang}, \bibinfo{person}{Megha Gupta},
  \bibinfo{person}{Michael Mahoney}, \bibinfo{person}{Priyank Goyal},
  \bibinfo{person}{Oleksandr Shchur}, \bibinfo{person}{Andrew Merrill},
  \bibinfo{person}{Caner Turkmen}, \bibinfo{person}{Qingsong Wen},
  \bibinfo{person}{Vasudev Rajan}, \bibinfo{person}{Renqian Luo},
  \bibinfo{person}{Christos Faloutsos}, {and} \bibinfo{person}{Yuyang Wang}.}
  \bibinfo{year}{2025}\natexlab{}.
\newblock \showarticletitle{{Mitra}: Mixed Synthetic Priors for Enhancing
  Tabular Foundation Models}. In \bibinfo{booktitle}{{\em Advances in Neural
  Information Processing Systems (NeurIPS)}}.
\newblock
\newblock
\shownote{arXiv:2510.21204.}


\bibitem[\protect\citeauthoryear{Zhang, Teng, Chowdhury, Li, Hong, Gupta, and
  Shang}{Zhang et~al\mbox{.}}{2024}]%
        {unimts}
\bibfield{author}{\bibinfo{person}{Xiyuan Zhang}, \bibinfo{person}{Diyan Teng},
  \bibinfo{person}{Ranak~Roy Chowdhury}, \bibinfo{person}{Shuheng Li},
  \bibinfo{person}{Dezhi Hong}, \bibinfo{person}{Rajesh~K. Gupta}, {and}
  \bibinfo{person}{Jingbo Shang}.} \bibinfo{year}{2024}\natexlab{}.
\newblock \showarticletitle{{UniMTS}: Unified Pre-training for Motion Time
  Series}. In \bibinfo{booktitle}{{\em Advances in Neural Information
  Processing Systems 37 (NeurIPS)}}.
\newblock


\bibitem[\protect\citeauthoryear{Zhaxidele, Bian, et~al\mbox{.}}{Zhaxidele
  et~al\mbox{.}}{2022}]%
        {recgym}
\bibfield{author}{\bibinfo{person}{Zhaxidele}, \bibinfo{person}{Jiawei Bian},
  {et~al\mbox{.}}} \bibinfo{year}{2022}\natexlab{}.
\newblock \bibinfo{title}{{RecGym}: Gym Workouts Recognition Dataset with {IMU}
  and Capacitive Sensor}.
\newblock \bibinfo{howpublished}{UCI Machine Learning Repository}.
  (\bibinfo{year}{2022}).
\newblock
\newblock
\shownote{\url{https://archive.ics.uci.edu/dataset/1128}.}


\end{thebibliography}

\end{document}